\documentclass{article}
\usepackage{arxiv,times}

\usepackage{amsmath}
\usepackage{amssymb}
\usepackage{graphicx}
\usepackage{booktabs}
\usepackage{multirow}
\usepackage{float}
\usepackage{wrapfig}
\usepackage{capt-of}
\usepackage[table]{xcolor}
\usepackage{colortbl}

\usepackage{hyperref}
\usepackage{url}

\title{$S^3$: Semantic Scale Synthesis for Training-Free Metric Depth from Complementary Foundation Models in Monocular 3DGS SLAM}

\author{Karthik Shaji, Zelin Zhao and Yongxin Chen \\
Georgia Institute of Technology \\
\texttt{kshaji3@gatech.edu}
}

\begin{document}

\maketitle

\begin{abstract}
Monocular depth foundation models can predict metric depth from individual images, but these predictions vary in scale and do not provide the multi-view geometric consistency required for estimation of motion while building a map. We introduce $S^3$-SLAM, a training-free monocular 3D Gaussian Splatting SLAM system that synthesizes two complementary foundation models: UniDepthV2, which infers single-frame metric depth, and VGGT, which infers multi-view-consistent but scale-free geometry. We show that their depth errors are complementary across spatial depth frequencies, introducing an online method of frequency-based fusion that exploits this structure while decoupling the mapping and tracking steps of SLAM. We use persistent semantic object landmarks to further correct frame-to-frame drift on short timescales, and to provide long-term geometric consistency when revisiting previously observed regions. By evaluating this approach across a range of indoor SLAM benchmarks, we recover near-metric-scale depth and trajectory estimates on most sequences, and identify the VGGT motion-degeneracy failure mode responsible for the sequences where tracking diverges.
\end{abstract}
\section{Introduction}
\label{sec:introduction}
\begin{wrapfigure}{r}{0.5\textwidth}
\centering
\vspace{-1.2em}
\includegraphics[width=\linewidth]{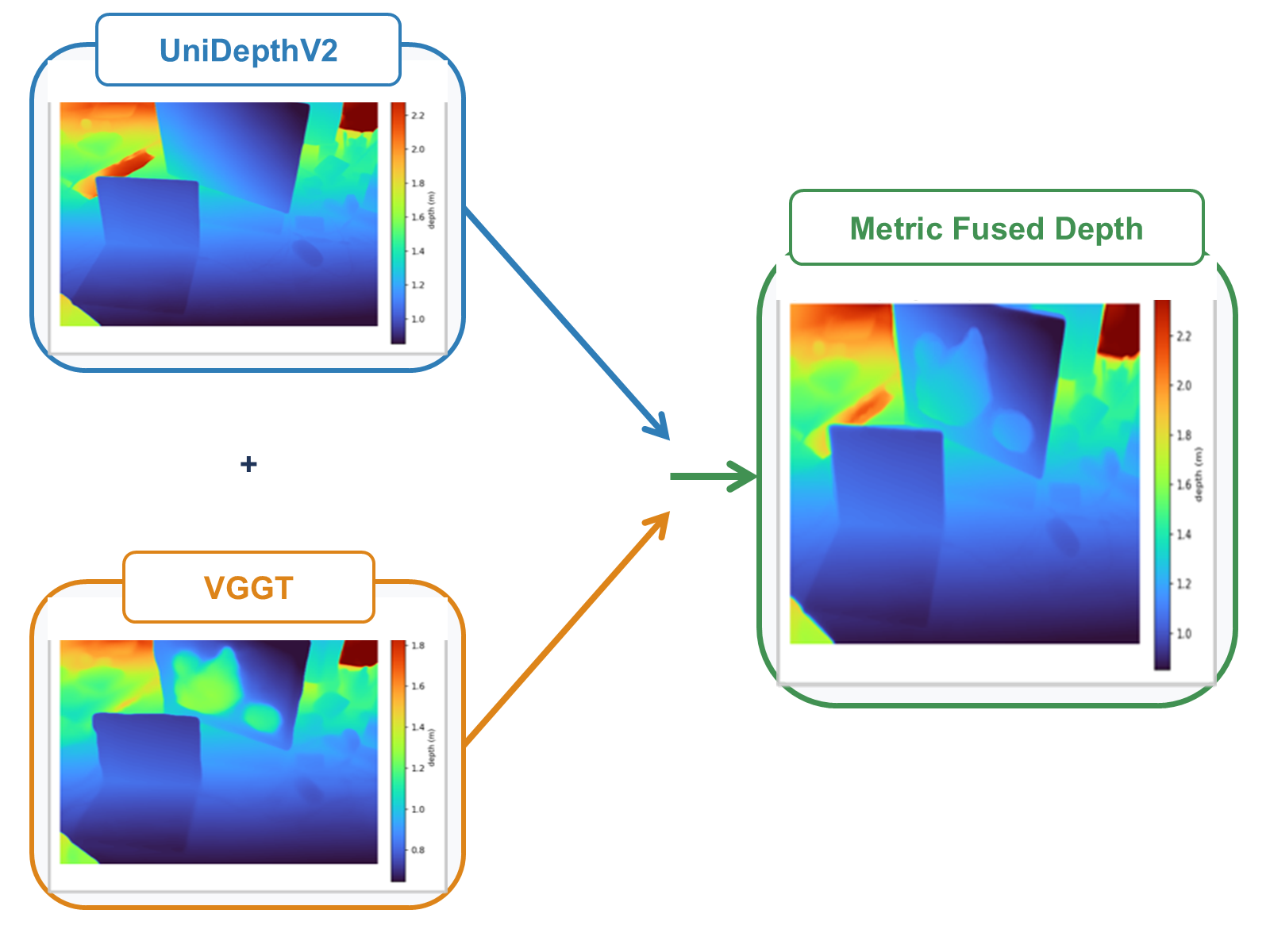}
\caption{\textbf{Complementary Depth Fusion.} Fusing two complementary foundation models yields geometrically-consistent metric depth.}
\label{fig:teaser}
\end{wrapfigure}

Monocular depth estimation aims to infer 3D scene structure from an individual image, but for a robot, this geometry must remain consistent across multiple sequential observations as the camera moves through the environment. Maintaining that consistency requires jointly reasoning about both the robot's motion and the structure of its environment. Simultaneous Localization and Mapping (SLAM) \citep{Leonard91tra_beacons} addresses this challenge by estimating camera motion while \emph{simultaneously} building a map of its surroundings. This process is critical because it enables a robot to maintain a spatially consistent map of an unknown environment while simultaneously tracking its own motion, making it fundamental for autonomous robot navigation \citep{Weiss11JFR_monoslam_nav}. The resulting pose estimates and maps support downstream tasks including navigation, trajectory optimization \citep{Lim26ral_monomorp} and collision avoidance \citep{Tscholl26ArXiv_FastBridge}. Without metric scale, the map can be geometrically consistent yet fail to tell the robot how far away an obstacle actually is. 

Recent monocular depth foundation models (DFMs) provide sources of metric information by predicting metric depth maps from single images \citep{Piccinelli26tpami_unidepthv2}. However, directly incorporating these maps into SLAM is not straightforward due to variation in scale or structure across subsequent observations. SLAM additionally requires geometric consistency over a continuous camera trajectory, due to how observations from different viewpoints must be integrated into the same map. While the Visual Geometry Grounded Transformer (VGGT) model \citep{Wang25cvpr_vggt} provides multi-view-consistent geometry and correspondence and has been used previously in SLAM \citep{Maggio25neurips_vggtslam}, it does not provide absolute metric scale. This motivates the problem of online synthesis of single-frame metric-level DFMs with multi-view-consistent, scale-free geometry foundation models (GFMs).

The distinction between scale-free and metric-level depth information becomes particularly important in vision-based SLAM systems that use 3D Gaussian Splatting (3DGS) \citep{Kerbl23tog_3dgs, Tosi26tro_nerfgsslamsurvey}. 3DGS uses compact Gaussian primitives to explicitly represent a scene while providing a continuous and differentiable map representation \citep{Kerbl23tog_3dgs}. In differentiable 3DGS SLAM \citep{Keetha24cvpr_splatam}, Gaussians and camera poses can be optimized online through rendering losses. As a result, errors in depth supervision not only affect an individual frame, but can also influence persistent map geometry and subsequent pose estimation during trajectory tracking. Existing monocular 3DGS SLAM approaches therefore commonly rely on photometric and relative geometric consistency \citep{Matsuki24cvpr_gsslam, Yu25icra_outdoorgs, Cheng25arxiv_outdoorgs}, rather than directly maintaining metrically consistent depth throughout mapping.

We address this challenge with 3DGS SLAM by decoupling how the mapping and pose estimation components of SLAM consume depth information. The mapping component fuses metric depth from UniDepthV2 with geometrically consistent depth from VGGT through an online optimization process that estimates reliability of the two depth sources. Simultaneously, the tracking component uses a scale-free version of this same depth map, enabling pose estimation to remain geometrically consistent while allowing the map to retain metric scale.

Another mechanism by which we improve geometric consistency in our map is through semantic object landmarks, which are identified through Segment Anything Model 2 (SAM2) \citep{Ravi24arxiv_sam2}. Prior work has used semantic landmarks to improve both feature correspondence \citep{Gan23TRO_multi} and map consistency \citep{Kim25UR_25}, as the identity of an object remains stable across changes in viewpoint. Once these landmarks are identified, we accumulate observations of them within the map over frames, which we can then compare against the current depth map, and use to further improve consistency via drift correction.

Integrating these mechanisms together in a semantic monocular RGB 3DGS SLAM system, $S^3$\emph{-SLAM}, our contributions are fourfold:
\begin{itemize}
\item We present an online, training-free fusion of two complementary monocular foundation models: VGGT's multi-view-consistent but scale-free geometry and UniDepthV2's metric single-frame depth.
\item We decouple the mapping and tracking, and introduce a method of online reliability optimization through spatial depth frequency analysis.
\item We show that persistent semantic object landmarks provide an additional online signal for correcting frame-dependent metric-depth scale error, while also supporting long-term drift correction.
\item We evaluate metric-level localization and mapping performance across TUM RGB-D \citep{Sturm12iros_tumrgbd}, 7-Scenes \citep{Glocker13ISMAR_realtime}, and ScanNet \citep{Dai2017ScanNet}.
\end{itemize}
\section{Related Work}
\label{sec:related_work}
\textbf{Scene Representations: } Dense scene representations encode geometry, color, and appearance throughout the observed environment. Neural Radiance Fields \citep{Mildenhall21cacm_nerf}, a neural-implicit method, use neural networks to predict density and color along a ray, but this approach suffers from a high computational cost due to the large number of MLP evaluations needed \citep{Huang24cvpr_photoslam}. 3DGS mitigates this issue through its explicit representation and avoidance of repeated neural-field evaluations at rendering time \citep{Tosi26tro_nerfgsslamsurvey, zhang2023goslamglobaloptimizationconsistent, yan2024gsslamdensevisualslam}. Our method uses 3DGS because it is a method that can be updated online while directly incorporating and optimizing depth. 

\textbf{Monocular SLAM: } Monocular SLAM estimates camera motion and map structure from a single RGB camera, causing metric scale to be ambiguous without additional information \citep{Davison07tpami_monoslam}. The ORB-SLAM algorithm, introduced by \citet{MurArtal15tro_orbslam} is a well-known feature-based algorithm which uses Oriented FAST and Rotated BRIEF (ORB) features \citep{Rublee11iccv_orb} as part of the optimization process, obtaining real-time performance without GPUs. While recent monocular 3DGS algorithms \citep{Matsuki24cvpr_gsslam, Zhang25_Hi} primarily optimize relative geometric and photometric consistency, and others \citep{Li26RAL_SING3R} use reconstruction priors for geometric consistency, our approach focuses on maintaining metric-depth consistency during online optimization. ConfidentSplat \citep{Dufera25_confidentsplat} and MyGO-Splat \citep{Zhu26_mygosplat} fuse depth via per-pixel confidence and closed-loop geometric feedback, respectively. We fuse per spatial-frequency band instead, using persistent semantic object identity rather than pixel confidence for scale correction, and our own confidence-weighted baseline is dataset-dependent, best on TUM and worst on ScanNet (Appendix~\ref{sec:appendix_fusion_mechanism_datasets}).

\textbf{Semantic SLAM: } Semantic SLAM utilizes object or category-level information in localization and mapping, rather than treating segmentation as a separate perception output \citep{Fu23Semantics}. Prior work has used semantic information to improve pose estimation, map consistency, and object-level mapping~\citep{Gan17arXiv_SparseBayes, Zhang18Access_ogsemslam}. Our method instead uses semantics to provide persistent object-level references for the correction of temporal metric-depth variation.
\section{Background}
\label{sec:background}
\subsection{3D Gaussian Splatting}
\label{sec:3dgs_background}

Introduced by \citet{Kerbl23tog_3dgs}, 3D Gaussian Splatting (3DGS) takes a set of images of a static scene, with corresponding cameras calibrated by Structure from Motion (SfM), and optimizes a set of Gaussian primitives to represent the scene. We follow several of the 3DGS methods used by \citet{Keetha24cvpr_splatam} when integrated into a SLAM pipeline: namely the silhouette-based optimization and the isotropic Gaussian formulation with view-independent color, where each Gaussian is parameterized by a 3D center, scalar radius, color, and opacity. The key equations and components of this method are given in Appendix~\ref{sec:appendix_3DGS}.

\subsection{Visual Geometry Grounded Transformer (VGGT)}
\label{sec:vggt_background}

VGGT, introduced by \citet{Wang25cvpr_vggt}, is a large feed-forward transformer that takes in a series of images as an input with the first one assumed to be the base reference frame (more formalism in \ref{appendix_vggt_formalism}). It predicts camera parameters, depth maps, point maps, and a grid of features. The translation and depth estimates are represented in a shared scene coordinate system without metric scale, following Eq.~\eqref{vggt_scale_formula}
\begin{equation}\label{vggt_scale_formula}
\begin{aligned}
d_{\mathrm{im}} &= d_{\mathrm{vggt}} \cdot \gamma_{\mathrm{vggt}}, \
m_{\mathrm{im}} = m_{\mathrm{vggt}} \cdot \gamma_{\mathrm{vggt}}.
\end{aligned}
\end{equation}
Here, $d$ is depth, $m$ is translation, and $\gamma_{\mathrm{vggt}}$ is the unknown scale factor mapping VGGT's internal scene scale to metric units; this scale factor is not output during inference. Thus, VGGT provides multi-view-consistent depth and translation estimates, while leaving their shared metric scale undetermined.

\subsection{Factor Graph Optimization}
\label{sec:factor_graph}
Factor graphs have been widely used in robotics and perception for SLAM and estimation \citep{Dellaert17book_factorgraphs}. We lay out our trajectory estimate as a nonlinear factor graph over camera poses $X = \{x_0, \dots, x_t\}$, where each $x_t$ is a camera pose in $SE(3)$, and solve it incrementally using incremental Smoothing and Mapping (iSAM2) \citep{Kaess11icra_isam2relin}, allowing us to fuse incoming pose estimates from VGGT in an online manner. We provide further background on factor graphs and iSAM2 in Appendix~\ref{sec:appendix_factor_graphs}.
\section{Monocular SLAM Architecture}
\label{sec:method}
\subsection{SLAM Terminology}
\label{sec:terminology}
We define several SLAM-specific terms that will be used in this section. First, we define a \emph{keyframe} as a subset of frames where computationally-expensive operations (such as VGGT inference) are called. While several classical SLAM algorithms \citep{MurArtal15tro_orbslam} determine them based on parallax or covisibility, we selected them at a fixed cadence (i.e. every $n$-th frame). \emph{Loop closure} is a mechanism by which drift in pose estimation is corrected by identifying when a robot has reached a previously visited point \citep{Hess16ICRA_LC}. Several strategies can be used for correcting this drift, including the breaking down of a global map into locally constructed chunks called \emph{submaps} which are then re-aligned \citep{Maggio26arxiv_vggtslam2, Zhu25_3dv_loopsplat}. The \emph{tracking} component of SLAM refers to the estimation of the camera pose of incoming frames relative to the existing map and previous observations.
\subsection{Algorithm Architecture Overview}
\begin{figure}
\centering
\includegraphics[width=0.7\linewidth]{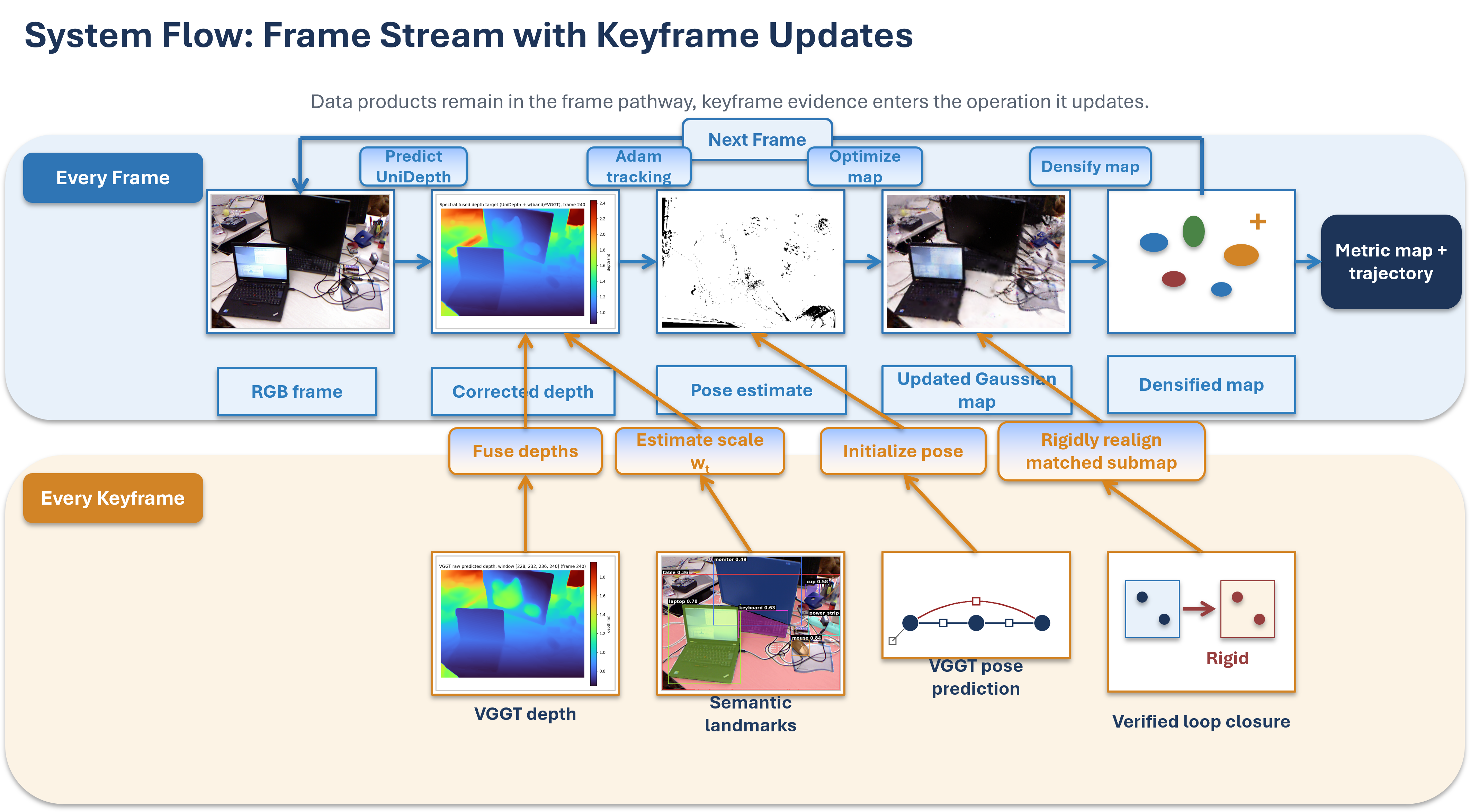}
\caption{{\bf Algorithm Overview.} The architecture of our algorithm is broken down into four distinct parts: depth fusion, tracking, loop closure detection, and Gaussian densification.}
\label{fig:alg_flow}
\end{figure}
We adopt the isotropic Gaussian and silhouette-based optimization of the RGB-D 3DGS SLAM system SplaTAM \citep{Keetha24cvpr_splatam} (given in Appendix~\ref{sec:appendix_3DGS}), but introduce four methods to maintain metric and geometric consistency over long monocular trajectories. They address four key estimation challenges: (1) combining single-view metric and multi-view-consistent scale-free predictions, (2) correcting frame-dependent variation from foundation model predictions, (3) recovering geometric consistency via loop closures, and (4) using VGGT to anchor pose estimates. A flowchart of the full system is shown in Fig.~\ref{fig:alg_flow}.

\subsection{Online Depth Fusion Via Spatial Depth Frequencies}
\label{sec:depth_fusion}
\begin{figure}
\centering
\includegraphics[width=0.7\linewidth]{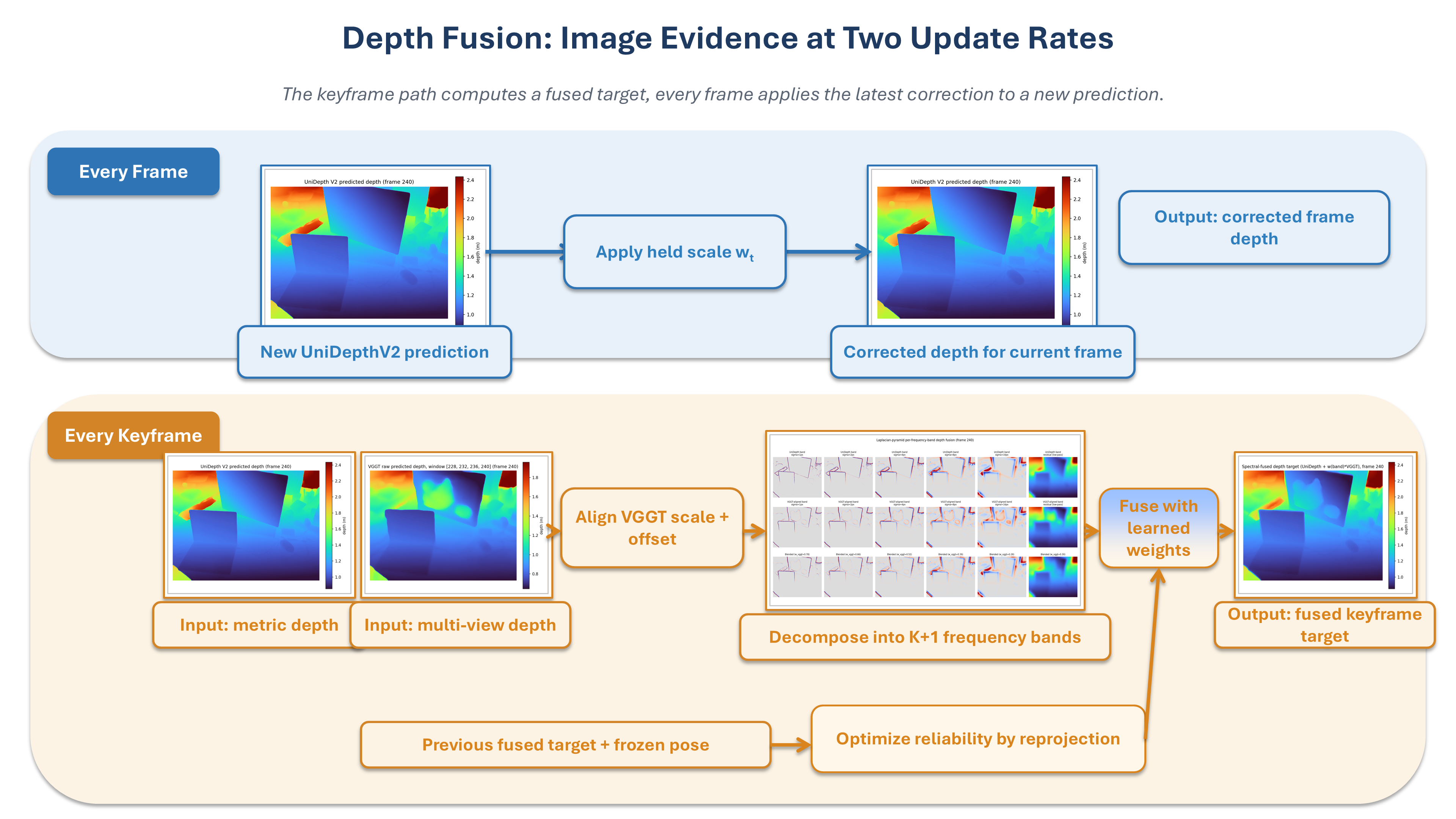}
\caption{{\bf Depth Fusion.} Our algorithm breaks down the spatial depth frequencies of VGGT and UniDepthV2 to produce a fused depth target for the Gaussian Splatting optimizer.}
\label{fig:depth_fusion}
\end{figure}
For each incoming frame, UniDepthV2 provides a single-frame metric-level depth estimate that can be used by the SLAM system; see Appendix~\ref{appendix_map_track_dense}. At every keyframe $\text{kf}_t$, we additionally run VGGT inference over a sliding window of the most recent keyframes up to $\text{kf}_{t-1}$, producing multi-view, scale-free depth estimates. We then perform depth fusion, which consists of three sequential components: (1) metric alignment of VGGT, (2) computation of cross-model reliability, and (3) frequency-dependent depth fusion.

VGGT infers multi-view-consistent depth maps that are scale-free, so we align this prediction to the current UniDepthV2 metric prediction using ~\eqref{depth_inv_eq} by jointly fitting a scale factor $\alpha^*$ and affine correction $b^*$.
\begin{equation}\label{depth_inv_eq}
(\alpha^{*}, b^{*}) = \arg\min_{\alpha, b}\left\| \mathbf{d}_{\text{uni}} - \left( \alpha \cdot \mathbf{d}_{\text{vggt}} + b \right) \right\|_2^2
\end{equation}
However, a single global scale factor is insufficient because the two models' depth errors differ across spatial frequency. We empirically observe that UniDepthV2 has lower error at coarse spatial scales, while VGGT has lower error at fine spatial scales due to multi-view triangulation. Let us define \emph{Spatial Depth Frequency} as the spatial rate of variation of a predicted depth field over image coordinates, and let $B_k(\mathbf{d})$ denote the band-pass component of depth field $\mathbf{d}$ at scale $k$, obtained by differencing successive Gaussian blurs of $\mathbf{d}$ over $N_b$ increasing scales (we use $N_b=7$), with $B_{N_b+1}$ the residual low-pass band. Our method computes cross-model reliability at each keyframe using a spatial depth frequency-dependent weight vector $\mathbf{w} = (w_1, \dots, w_{N_b+1})$. Given the aligned VGGT depth $\hat{\mathbf{d}}_{\text{vggt}}$ from Eq.~\eqref{depth_inv_eq}, the fused depth target is
\begin{equation}\label{spectral_fuse_eq}
\mathbf{d}_{\text{fused}}(\mathbf{w})
= \sum_{k=1}^{N_b+1} \Big[ B_k(\mathbf{d}_{\text{uni}})
+ w_k \big( B_k(\hat{\mathbf{d}}_{\text{vggt}}) - B_k(\mathbf{d}_{\text{uni}}) \big) \Big].
\end{equation}
$w_k=0$ selects UniDepthV2 for band $k$ while $w_k=1$ selects aligned VGGT, with intermediate values interpolating between the two. We estimate $\mathbf{w}$ by using the tracker's frozen pose estimate to reproject $\mathbf{d}_{\text{fused}}(\mathbf{w})$ into the previous keyframe's view and compare it against the previous committed fused depth. Before measuring this residual, we remove the residual global scale and offset mismatch via an auxiliary affine fit, so that a uniform scale error is not mistaken for depth-dependent shape error. We regularize $\mathbf{w}$ towards the previous keyframe's weight vector $\mathbf{w}_{t-1}$ so that band weights vary smoothly across keyframes, while still constraining them to $[0, 1]$:
\begin{equation}
\label{eq:spectral_band_weight}
\mathbf{w}_t = \arg\min_{\mathbf{w} \in [0,1]^{N_b+1}} \; L_{\text{geo}}(\mathbf{w}) + \lambda \left\| \mathbf{w} - \mathbf{w}_{t-1} \right\|_2^2.
\end{equation}
In our implementation, we solve this optimization via zeroth-order coordinate search over $L_{\text{geo}}$, without a photometric term.
\subsection{Semantic Metric-Level Correction}
\label{sec:semantic_scale}
\begin{figure}
\centering
\includegraphics[width=0.7\linewidth]{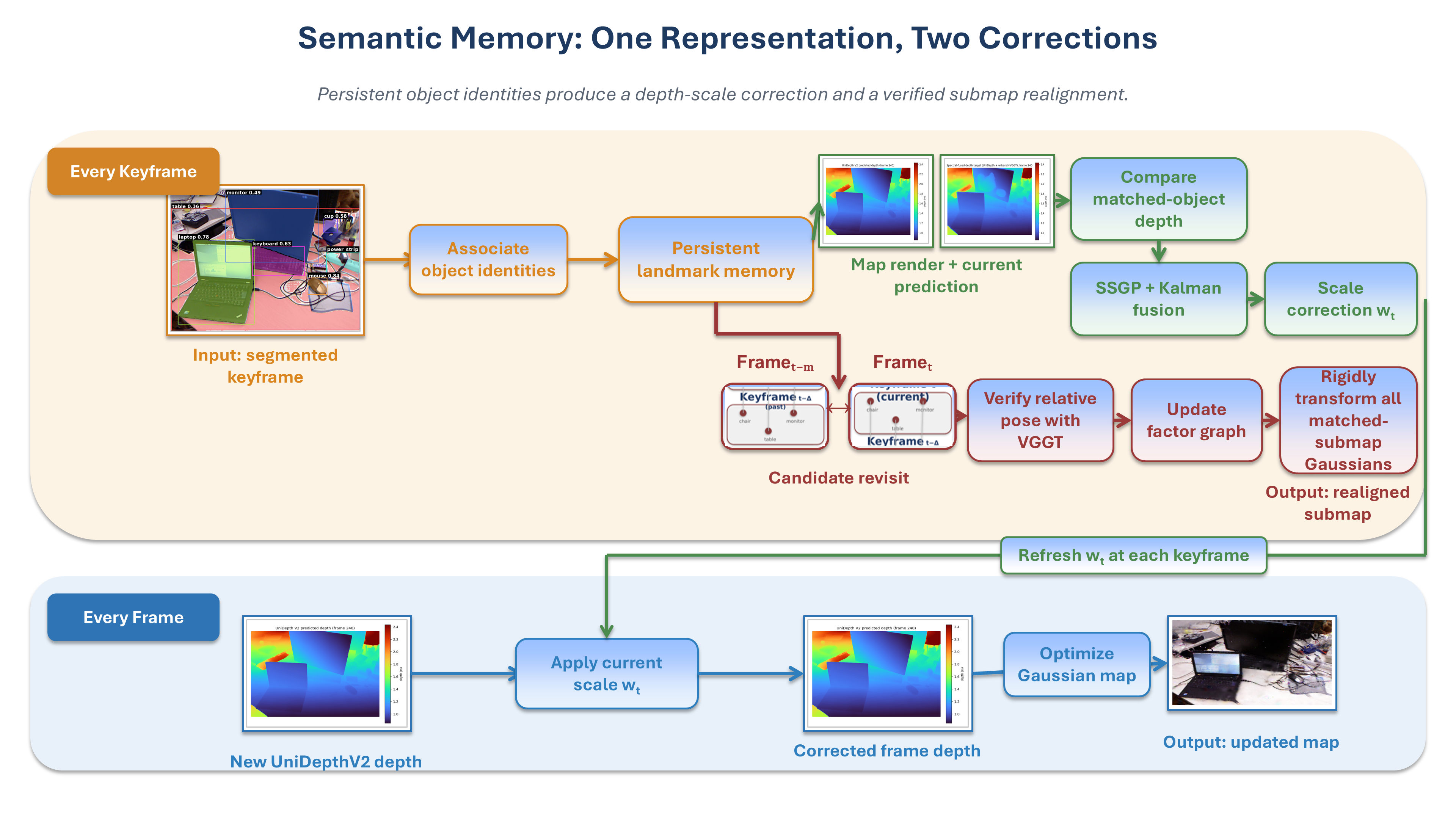}
\caption{{\bf Semantic Analysis Overview.} The semantics are used in two ways: to provide more robust depth estimation, and for loop closure/rigid submap corrections}
\label{sem_flow}
\end{figure}
Depth fusion addresses differences in spatial structure between UniDepthV2 and VGGT, but it does not address frame-dependent scale variation in UniDepthV2's single-frame predictions. We therefore use objects already accumulated in the map as persistent references for estimating temporal deviations in metric scale.

Obtaining this signal requires persistent object correspondences rather than raw semantic labels. At each keyframe $\text{kf}_t$, we apply Grounding DINO \citep{Liu24eccv_groundingdino}, followed by box-prompted SAM2 \citep{Ravi24arxiv_sam2}, to obtain scored instance masks. We associate instances across keyframes using VGGT point-track correspondences and a lightweight image-space Kalman tracker. A chi-square innovation gate rejects ambiguous associations between objects with the same semantic label.

Each Gaussian inserted between keyframes $\text{kf}_{t-1}$ and $\text{kf}_t$ inherits the identity of the track-validated instance containing its birth pixel. A semantic landmark is therefore represented by a subset of the map's Gaussians rather than a separate geometric state. Gaussian additions are grouped by keyframe, and inconsistent groups are rejected before the remaining observations are combined. For each visible mature landmark, we compute the median ratio between its rendered depth and the current UniDepthV2 prediction. This ratio measures frame-dependent scale variation relative to the accumulated map, and semantics provide the correspondence needed to construct the measurement, rather than an independent source of metric depth.

The accumulated map also contains uncertainty from previous depth and pose estimates, so landmark ratios cannot be applied directly. We assign each category a measurement variance based on the number and agreement of its landmark observations:
\begin{equation}\label{landmark_meas}
\sigma^2_{\text{meas},i}
= \frac{\kappa^2 \operatorname{MMAD}_i^2}{n_i}
+ \sigma_{\text{floor}}^2,
\end{equation}
where $\operatorname{MMAD}_i$ denotes the mean median absolute deviation for category $i$, $\kappa$ scales this dispersion, and $\sigma_{\text{floor}}^2$ is a noise floor. This gives inconsistent categories less weight. To avoid pooling category-dependent errors, we track each category's log-ratio with a state-space Gaussian process using $\sigma_{\text{meas},i}^2$ as its measurement noise. Its state $\mathbf{x}_t$, containing Ornstein--Uhlenbeck and Matérn-3/2 components, is estimated at each keyframe by
\begin{equation}
\mathbf{x}_t = \arg\min_{\mathbf{x}}
(\mathbf{x} - A\mathbf{x}_{t-1})^\top
\Sigma_t^{-1}
(\mathbf{x} - A\mathbf{x}_{t-1})
+ \frac{(y_t - H\mathbf{x})^2}{R_t},
\end{equation}
where $A$ and $H$ are derived from the underlying Matérn kernels and $\Sigma_t = AP_{t-1}A^\top + \mathbf{Q}_{\mathrm{GP}}$. The resulting category estimates and their uncertainties are fused with a scalar Kalman filter over $\delta$ (the log-scale correction):
\begin{equation}
\hat{\delta}_t = \arg\min_{\delta}
\frac{(\delta-\hat{\delta}_{t-1})^2}{P_{t-1}+q_\delta}
+ \frac{(z_t-\delta)^2}{R_t},
\end{equation}
where $z_t$ is the fused measurement across visible categories and $R_t$ is its variance. The semantic correction $\eta_t = \exp(\hat{\delta}_t)$ is estimated at keyframes, then applied to the UniDepthV2 depth predictions of subsequent frames until the next keyframe update, with a gradual decay toward $1$ between updates. Appendix~\ref{sec:appendix_semantic_tracking} provides the complete filter updates.
\subsection{Semantic Loop Closure}
\label{sec:loop_closure}
At each keyframe after semantic landmarks have been updated, the persistent object-level representation gives us a pose-independent mechanism for identifying loop closures. Each semantic landmark has a corrected depth, so a revisited location stays consistent even after the estimated pose has drifted. Once semantic agreement, obtained through label consistency and pairwise correspondence, identifies candidates, we use VGGT to estimate a relative pose between the current keyframe and the previously observed keyframe. We then verify each candidate geometrically, accepting the candidate only if VGGT triangulates the pair with sufficient confidence and its relative pose prediction disagrees with the current pose estimate by more than a minimum drift threshold.

When a loop closure is verified, we apply its pose correction to the previously constructed Gaussian submap for that region, rather than to only the current frame, similar to \citep{Zhu25_3dv_loopsplat}. Each submap's anchor frame is $a_j = j \cdot \texttt{submap\_size}$, where $j$ is the submap index and \texttt{submap\_size} is a user-set parameter. The correction is a single rigid transform between that anchor's extrinsic matrix before and after the factor-graph update,
\begin{equation}
T_j^{\mathrm{loop}} = E^{\text{new}}_{a_j} \left(E^{\text{old}}_{a_j}\right)^{-1},
\end{equation}
which is applied to every Gaussian in the submap. Subsequent tracking and mapping then use this corrected submap geometry. We show the overall methodology of how our algorithm uses semantics in Fig.~\ref{sem_flow}.
\subsection{Pose Estimation with VGGT}
\label{sec:vggt_pose}
VGGT provides pose information only at keyframes, while Adam tracking estimates camera motion for every incoming frame. At each keyframe, we initialize the Adam pose optimization from this VGGT prediction rather than extrapolating the previous two poses. We only use it when it agrees with the extrapolated estimate within a fixed translation and rotation bound. Otherwise we  
fall back to the extrapolated estimate, since VGGT occasionally produces implausible predictions. Once tracking converges, we also add VGGT's measured relative pose between
the two most recent keyframes in the window as an extra edge in the factor graph (Sec.~\ref{sec:appendix_factor_graphs}), alongside the edge connecting every consecutive   
pair of frames from Adam's tracked motion. The pose graph then solves for the pose estimate using iSAM2.
\section{Experimental Results}
\label{sec:result}

\newcommand{\tablebodyfont}{\fontsize{6}{7.2}\selectfont}

\newcommand{\TumRenderingTable}{
\begin{table}[H]
\centering
\caption{{\bf Rendering performance on TUM RGB-D.} Keyframes only, following Splat-SLAM's protocol. Baselines as reported. Best \colorbox{green!25}{1st}, \colorbox{yellow!25}{2nd}, \colorbox{orange!25}{3rd}.}
\label{tab:tum_rendering_splatslam}
\tablebodyfont
\setlength{\tabcolsep}{4pt}
\begin{tabular}{ll cccc c}
\toprule
\textbf{Method} & \textbf{Metric} & \textbf{\emph{fr1/desk}} & \textbf{\emph{fr2/xyz}} & \textbf{\emph{fr3/office}} & \textbf{\emph{fr1/desk2}} & \textbf{\emph{fr1/room}} \\
\midrule
Photo-SLAM
 & PSNR$\uparrow$ & \cellcolor{yellow!25}20.97 & \cellcolor{orange!25}21.07 & 19.59 & -- & -- \\
 & SSIM$\uparrow$ & -- & -- & -- & -- & -- \\
 & LPIPS$\downarrow$ & \cellcolor{yellow!25}0.23 & \cellcolor{orange!25}0.17 & \cellcolor{yellow!25}0.24 & -- & -- \\
MonoGS
 & PSNR$\uparrow$ & 19.67 & 16.17 & \cellcolor{orange!25}20.63 & \cellcolor{orange!25}19.16 & 18.41 \\
 & SSIM$\uparrow$ & \cellcolor{orange!25}0.73 & \cellcolor{yellow!25}0.72 & \cellcolor{yellow!25}0.77 & 0.66 & 0.64 \\
 & LPIPS$\downarrow$ & 0.33 & 0.31 & 0.34 & 0.48 & 0.51 \\
GlORIE-SLAM
 & PSNR$\uparrow$ & \cellcolor{orange!25}20.26 & \cellcolor{yellow!25}25.62 & \cellcolor{yellow!25}21.21 & 19.09 & \cellcolor{orange!25}18.78 \\
 & SSIM$\uparrow$ & \cellcolor{yellow!25}0.79 & \cellcolor{yellow!25}0.72 & \cellcolor{orange!25}0.72 & \cellcolor{green!25}0.92 & \cellcolor{yellow!25}0.73 \\
 & LPIPS$\downarrow$ & \cellcolor{orange!25}0.31 & \cellcolor{yellow!25}0.09 & \cellcolor{orange!25}0.32 & \cellcolor{yellow!25}0.38 & \cellcolor{orange!25}0.38 \\
Splat-SLAM
 & PSNR$\uparrow$ & \cellcolor{green!25}25.61 & \cellcolor{green!25}29.53 & \cellcolor{green!25}26.05 & \cellcolor{green!25}23.98 & \cellcolor{green!25}24.06 \\
 & SSIM$\uparrow$ & \cellcolor{green!25}0.84 & \cellcolor{green!25}0.90 & \cellcolor{green!25}0.84 & \cellcolor{yellow!25}0.81 & \cellcolor{green!25}0.80 \\
 & LPIPS$\downarrow$ & \cellcolor{green!25}0.18 & \cellcolor{green!25}0.08 & \cellcolor{green!25}0.20 & \cellcolor{green!25}0.23 & \cellcolor{green!25}0.24 \\
\textbf{Ours}
 & PSNR$\uparrow$ & 18.92 & 19.27 & 18.97 & \cellcolor{yellow!25}19.44 & \cellcolor{yellow!25}19.34 \\
 & SSIM$\uparrow$ & 0.701 & 0.665 & 0.690 & \cellcolor{orange!25}0.719 & \cellcolor{orange!25}0.698 \\
 & LPIPS$\downarrow$ & 0.398 & 0.402 & 0.470 & \cellcolor{orange!25}0.396 & \cellcolor{yellow!25}0.376 \\
\bottomrule
\end{tabular}
\end{table}
}

\newcommand{\TumDepthLOneTable}{
\begin{table}[H]
\centering
\caption{{\bf Depth L1 error on TUM RGB-D.}}
\label{tab:tum_depth_l1}
\tablebodyfont
\setlength{\tabcolsep}{7pt}
\renewcommand{\arraystretch}{1.15}
\begin{tabular}{|l|c|c|c|c|c|c|}
\hline
\textbf{Scene}
& \textbf{\emph{fr1/desk}}
& \textbf{\emph{fr2/xyz}}
& \textbf{\emph{fr3/office}}
& \textbf{\emph{fr1/desk2}}
& \textbf{\emph{fr1/room}}
& \textbf{Avg.} \\
\hline
\textbf{Depth L1 [cm]}
& 8.29
& 19.67
& 28.92
& 12.84
& 13.14
& \textbf{16.57} \\
\hline
\end{tabular}
\end{table}
}

\newcommand{\SevenScenesTable}{
\begin{table}[H]
\centering
\caption{{\bf Sim(3) ATE RMSE [cm] on 7-Scenes.} Following the evaluation protocol of \citep{Li26RAL_SING3R}. Best \colorbox{green!25}{1st}, \colorbox{yellow!25}{2nd}, \colorbox{orange!25}{3rd}.}
\label{tab:sevenscenes_results}
\tablebodyfont
\setlength{\tabcolsep}{5pt}
\renewcommand{\arraystretch}{1.15}
\begin{tabular}{|l|c|c|c|c|c|c|c|}
\hline
\textbf{Method} & \textbf{\emph{chess}} & \textbf{\emph{fire}} & \textbf{\emph{heads}} & \textbf{\emph{office}} & \textbf{\emph{pumpkin}} & \textbf{\emph{redkitchen}} & \textbf{\emph{stairs}} \\
\hline
VGGT-SLAM
& \cellcolor{yellow!25}3.6
& \cellcolor{yellow!25}2.8
& \cellcolor{green!25}1.8
& 10.3
& \cellcolor{yellow!25}13.3
& 5.8
& 9.3 \\
\hline
HI-SLAM2
& \cellcolor{orange!25}3.8
& \cellcolor{orange!25}3.1
& \cellcolor{yellow!25}2.6
& \cellcolor{orange!25}8.5
& \cellcolor{orange!25}14.2
& \cellcolor{yellow!25}4.0
& \cellcolor{green!25}2.4 \\
\hline
SING3R-SLAM
& \cellcolor{green!25}3.3
& \cellcolor{green!25}2.7
& 4.5
& \cellcolor{green!25}7.2
& \cellcolor{green!25}8.7
& \cellcolor{green!25}3.6
& \cellcolor{orange!25}4.5 \\
\hline
Ours
& 5.72
& 3.70
& \cellcolor{orange!25}3.90
& \cellcolor{yellow!25}8.45
& 16.01
& \cellcolor{orange!25}5.2
& \cellcolor{yellow!25}4.1 \\
\hline
\end{tabular}
\end{table}
}

\newcommand{\LoopClosureTable}{
\begin{table}[H]
\centering
\caption{{\bf Loop closure precision and recall.} On TUM \emph{fr1/desk} and \emph{fr1/desk2}. Best \colorbox{green!25}{1st}, \colorbox{yellow!25}{2nd}.}
\label{tab:lc_precision_recall}
\tablebodyfont
\setlength{\tabcolsep}{6pt}
\begin{tabular}{lccc}
\toprule
Method
& Recall $\uparrow$
& TP density $\uparrow$
& Precision $\uparrow$ \\
\midrule
Ours (semantics on)
& \cellcolor{green!25}0.500
& \cellcolor{green!25}6.00
& \cellcolor{yellow!25}0.571 \\

No semantics (pose-proximity only)
& \cellcolor{yellow!25}0.250
& \cellcolor{yellow!25}2.00
& \cellcolor{green!25}1.000\\
\bottomrule
\end{tabular}

\vspace{2pt}
\end{table}
}

\newcommand{\AblationsTable}{
\begin{table}[H]
\centering
\caption{{\bf Ablation study.} TUM RGB-D \emph{fr1/desk2} sequence. Best \colorbox{green!25}{1st}, \colorbox{yellow!25}{2nd}, \colorbox{orange!25}{3rd}.}
\label{tab:ablations}
\tablebodyfont
\setlength{\tabcolsep}{5pt}
\renewcommand{\arraystretch}{1.05}
\begin{tabular}{lccccccc}
\toprule
Configuration
& Sim(3) ATE $\downarrow$
& SE(3) ATE $\downarrow$
& Scale $s$
& PSNR $\uparrow$
& SSIM $\uparrow$
& LPIPS $\downarrow$
& Depth L1 $\downarrow$ \\
\midrule

Full (Ours)
& \cellcolor{green!25}4.06
& \cellcolor{yellow!25}4.87
& 1.0267
& \cellcolor{green!25}19.44
& \cellcolor{green!25}0.719
& \cellcolor{green!25}0.396
& \cellcolor{green!25}12.84 \\

No semantics
& 10.60
& 15.21
& \cellcolor{yellow!25}1.0045
& 17.40
& 0.689
& 0.441
& 15.24 \\

\quad Scale correction only
& 9.39
& 14.04
& 0.9934
& 17.63
& \cellcolor{orange!25}0.694
& 0.431
& 14.83 \\

\quad LC proposals only
& \cellcolor{yellow!25}4.39
& \cellcolor{green!25}4.70
& 1.0067
& \cellcolor{orange!25}19.25
& \cellcolor{yellow!25}0.714
& \cellcolor{yellow!25}0.399
& \cellcolor{orange!25}13.57 \\

No depth fusion
& \cellcolor{orange!25}4.59
& \cellcolor{orange!25}4.94
& \cellcolor{green!25}1.0031
& \cellcolor{yellow!25}19.26
& \cellcolor{yellow!25}0.714
& \cellcolor{orange!25}0.402
& \cellcolor{yellow!25}13.33 \\

No loop closure
& 7.39
& 8.43
& \cellcolor{orange!25}0.9954
& 17.37
& 0.692
& 0.426
& 15.19 \\

\bottomrule
\end{tabular}
\end{table}
}

\newcommand{\FusionMechanismTable}{
\begin{table}[H]
\centering
\caption{{\bf Online fusion vs.\ a frozen, GT-fitted band-weight curve.} TUM RGB-D \emph{fr1/desk2}, same protocol, same seed. The frozen curve is fit once offline against ground truth and cannot be computed at inference time; it is shown only as an upper-bound reference for a fixed-curve alternative. The online optimizer is competitive with or behind the frozen reference on every metric except Sim(3) ATE, where it is the only one of the two with any advantage. Best \colorbox{green!25}{1st}.}
\label{tab:fusion_mechanism}
\tablebodyfont
\setlength{\tabcolsep}{5pt}
\renewcommand{\arraystretch}{1.05}
\begin{tabular}{lccccc}
\toprule
Fusion mechanism
& Sim(3) ATE $\downarrow$
& PSNR $\uparrow$
& SSIM $\uparrow$
& LPIPS $\downarrow$
& Depth L1 $\downarrow$ \\
\midrule

Online optimizer (Ours, shipped)
& \cellcolor{green!25}4.46
& 19.15
& 0.712
& 0.398
& 14.09 \\

Frozen GT-fitted curve
& 4.69
& \cellcolor{green!25}19.27
& \cellcolor{green!25}0.715
& \cellcolor{green!25}0.396
& \cellcolor{green!25}13.50 \\

\bottomrule
\end{tabular}
\end{table}
}

\newcommand{\ScanNetAblationsTable}{
\begin{table}[H]
\centering
\caption{{\bf Ablation study.} ScanNet scene \emph{0059}, full sequence. ``Generic correspondence'' replaces semantic object/plane landmarks with a non-semantic persistent-correspondence tracker built on the same grid-based point tracking VGGT already computes, grouped by fixed image-space cell rather than semantic instance, feeding the identical downstream scale-correction and loop-closure pipeline. Best \colorbox{green!25}{1st}, \colorbox{yellow!25}{2nd}, \colorbox{orange!25}{3rd}.}
\label{tab:scannet_ablations}
\tablebodyfont
\setlength{\tabcolsep}{5pt}
\renewcommand{\arraystretch}{1.05}
\begin{tabular}{lccccc}
\toprule
Configuration
& Sim(3) ATE $\downarrow$
& PSNR $\uparrow$
& SSIM $\uparrow$
& LPIPS $\downarrow$
& Depth RMSE $\downarrow$ \\
\midrule

Full (Ours)
& \cellcolor{green!25}12.03
& \cellcolor{green!25}18.86
& \cellcolor{green!25}0.704
& \cellcolor{green!25}0.524
& \cellcolor{green!25}35.51 \\

No semantics
& 13.05
& 18.25
& 0.690
& \cellcolor{orange!25}0.543
& \cellcolor{yellow!25}37.37 \\

\quad Scale correction only
& \cellcolor{yellow!25}12.12
& 18.06
& 0.679
& 0.562
& 41.37 \\

\quad LC proposals only
& \cellcolor{orange!25}12.66
& \cellcolor{yellow!25}18.66
& \cellcolor{yellow!25}0.698
& \cellcolor{yellow!25}0.526
& \cellcolor{orange!25}37.46 \\

No depth fusion
& 12.91
& \cellcolor{orange!25}18.28
& 0.687
& 0.553
& 40.93 \\

No loop closure
& 35.65
& 17.99
& \cellcolor{orange!25}0.691
& 0.557
& 51.42 \\

Generic correspondence
& 13.78
& 18.05
& 0.681
& 0.557
& 39.61 \\

\bottomrule
\end{tabular}
\end{table}
}

\newcommand{\TumAteTable}{
\begin{table}[H]
\centering
\hspace*{0.03\linewidth}%
\begin{tabular}{@{}p{0.44\linewidth}@{\hspace{0.06\linewidth}}p{0.44\linewidth}@{}}
\begin{minipage}[t]{\linewidth}
\vspace{0pt}
\centering
\captionof{table}{{\bf TUM RGB-D ATE RMSE [cm].} SE(3) ATE compares to RGB-D baselines; Sim(3) compares to monocular baselines. Baseline values are from Splat-SLAM~\citep{Sandstrom25CVPRW_splatslam} and LoopSplat~\citep{Zhu25_3dv_loopsplat}. Our fitted scale $s$ for Sim(3) is reported below. Best \colorbox{green!25}{1st}, \colorbox{yellow!25}{2nd}, \colorbox{orange!25}{3rd}.}
\label{tab:tum_ate_results}
\end{minipage}
&
\begin{minipage}[t]{\linewidth}
\vspace{0pt}
\centering
\captionof{table}{{\bf ScanNet ATE RMSE [cm].} SE(3) ATE compares to RGB-D baselines; Sim(3) ATE compares to monocular baselines. Baseline values are from Splat-SLAM~\citep{Sandstrom25CVPRW_splatslam} and LoopSplat~\citep{Zhu25_3dv_loopsplat}. Our fitted scale $s$ for Sim(3) is reported below. Ours failed on scene \emph{0181}. Best \colorbox{green!25}{1st}, \colorbox{yellow!25}{2nd}, \colorbox{orange!25}{3rd}.}
\label{tab:scannet_ate_results}
\end{minipage}
\\
\begin{minipage}[t]{\linewidth}
\vspace{0pt}
\centering
\tablebodyfont
\setlength{\tabcolsep}{3pt}
\renewcommand{\arraystretch}{1.15}
\begin{tabular}{|l|c|c|c|c|c|}
\hline
\textbf{Method} & \textbf{\emph{fr1/desk}} & \textbf{\emph{fr2/xyz}} & \textbf{\emph{fr3/office}} & \textbf{\emph{fr1/desk2}} & \textbf{\emph{fr1/room}} \\
\hline
\multicolumn{6}{|l|}{\textit{SE(3), vs.\ RGB-D}} \\
\hline
SplaTAM & 3.4 & \cellcolor{yellow!25}1.2 & 5.2 & 6.5 & 11.1 \\
\hline
GS-SLAM & \cellcolor{yellow!25}1.5 & 1.6 & \cellcolor{orange!25}1.7 & -- & -- \\
\hline
GO-SLAM (RGB-D) & \cellcolor{yellow!25}1.5 & \cellcolor{green!25}0.6 & \cellcolor{green!25}1.3 & -- & \cellcolor{green!25}4.7 \\
\hline
MonoGS (RGB-D) & \cellcolor{green!25}1.4 & \cellcolor{orange!25}1.4 & \cellcolor{yellow!25}1.5 & \cellcolor{orange!25}5.1 & \cellcolor{orange!25}6.3 \\
\hline
LoopSplat & 2.08 & 1.58 & 3.22 & \cellcolor{green!25}3.54 & \cellcolor{yellow!25}6.24 \\
\hline
Ours (SE(3)) & 4.51 & 2.93 & 14.32 & \cellcolor{yellow!25}4.87 & 44.52 \\
\hline
\multicolumn{6}{|l|}{\textit{Sim(3), vs.\ monocular}} \\
\hline
Photo-SLAM & \cellcolor{green!25}1.5 & \cellcolor{yellow!25}1.0 & \cellcolor{green!25}1.3 & -- & -- \\
\hline
MonoGS & \cellcolor{orange!25}3.8 & 5.2 & \cellcolor{orange!25}2.9 & \cellcolor{orange!25}75.7 & \cellcolor{orange!25}76.6 \\
\hline
Splat-SLAM & \cellcolor{yellow!25}1.6 & \cellcolor{green!25}0.2 & \cellcolor{yellow!25}1.4 & \cellcolor{green!25}2.8 & \cellcolor{green!25}4.2 \\
\hline
Ours (Sim(3)) & 4.47 & \cellcolor{orange!25}2.93 & 6.64 & \cellcolor{yellow!25}4.06 & \cellcolor{yellow!25}41.04 \\
\hline
Ours: scale $s$ & 0.9926 & 1.0046 & 0.9535 & 1.0267 & 1.1027 \\
\hline
\end{tabular}
\end{minipage}
&
\begin{minipage}[t]{\linewidth}
\vspace{0pt}
\centering
\tablebodyfont
\setlength{\tabcolsep}{3pt}
\renewcommand{\arraystretch}{1.15}
\begin{tabular}{|l|c|c|c|c|c|c|}
\hline
\textbf{Method} & \textbf{\emph{0000}} & \textbf{\emph{0059}} & \textbf{\emph{0106}} & \textbf{\emph{0169}} & \textbf{\emph{0181}} & \textbf{\emph{0207}} \\
\hline
\multicolumn{7}{|l|}{\textit{SE(3), vs.\ RGB-D}} \\
\hline
SplaTAM & \cellcolor{yellow!25}12.8 & \cellcolor{orange!25}10.1 & 17.7 & \cellcolor{orange!25}12.1 & \cellcolor{yellow!25}11.1 & \cellcolor{yellow!25}7.5 \\
\hline
GS-SLAM & 21.2 & 12.8 & \cellcolor{orange!25}13.5 & 16.3 & \cellcolor{orange!25}21.0 & 14.3 \\
\hline
GO-SLAM (RGB-D) & -- & -- & -- & -- & -- & -- \\
\hline
MonoGS (RGB-D) & \cellcolor{orange!25}16.1 & \cellcolor{green!25}6.4 & \cellcolor{yellow!25}8.1 & \cellcolor{green!25}8.7 & 26.4 & \cellcolor{orange!25}9.2 \\
\hline
LoopSplat & \cellcolor{green!25}6.2 & \cellcolor{yellow!25}7.1 & \cellcolor{green!25}7.4 & \cellcolor{yellow!25}10.6 & \cellcolor{green!25}8.5 & \cellcolor{green!25}6.6 \\
\hline
Ours (SE(3)) & 171.87 & 16.34 & 16.93 & 30.17 & x & 222.46 \\
\hline
\multicolumn{7}{|l|}{\textit{Sim(3), vs.\ monocular}} \\
\hline
Photo-SLAM & -- & -- & -- & -- & -- & -- \\
\hline
MonoGS & \cellcolor{yellow!25}149.2 & \cellcolor{orange!25}96.8 & \cellcolor{orange!25}155.5 & \cellcolor{orange!25}140.3 & \cellcolor{yellow!25}92.6 & \cellcolor{yellow!25}101.9 \\
\hline
Splat-SLAM & \cellcolor{green!25}5.5 & \cellcolor{green!25}9.1 & \cellcolor{green!25}7.0 & \cellcolor{green!25}8.2 & \cellcolor{green!25}8.3 & \cellcolor{green!25}7.5 \\
\hline
Ours (Sim(3)) & \cellcolor{orange!25}160.07 & \cellcolor{yellow!25}16.23 & \cellcolor{yellow!25}15.76 & \cellcolor{yellow!25}29.29 & x & \cellcolor{orange!25}222.22 \\
\hline
Ours: scale $s$ & 1.2308 & 1.0147 & 1.022 & 0.953 & x & 1.0599 \\
\hline
\end{tabular}
\end{minipage}
\end{tabular}%
\hspace*{0.03\linewidth}
\end{table}
}

\newcommand{\ScanNetRenderingTable}{
\begin{table}[H]
\centering
\caption{{\bf Rendering performance on ScanNet.} Monocular baselines only, from \citep{Sandstrom25CVPRW_splatslam}. $^\dagger$Ours excludes scene \emph{0181} (tracking failed). Best \colorbox{green!25}{1st}, \colorbox{yellow!25}{2nd}, \colorbox{orange!25}{3rd}.}
\label{tab:scannet_rendering_results}
\tablebodyfont
\setlength{\tabcolsep}{3.5pt}
\begin{tabular}{ll cccccc c}
\toprule
\textbf{Method} & \textbf{Metric} & \textbf{\emph{0000}} & \textbf{\emph{0059}} & \textbf{\emph{0106}} & \textbf{\emph{0169}} & \textbf{\emph{0181}} & \textbf{\emph{0207}} & \textbf{Avg.} \\
\midrule
MonoGS
 & PSNR$\uparrow$ & 16.91 & \cellcolor{orange!25}19.15 & 18.57 & \cellcolor{orange!25}20.21 & \cellcolor{orange!25}19.51 & 18.37 & 18.79 \\
 & SSIM$\uparrow$ & 0.62 & 0.69 & \cellcolor{orange!25}0.74 & \cellcolor{orange!25}0.74 & \cellcolor{orange!25}0.75 & 0.70 & 0.71 \\
 & LPIPS$\downarrow$ & 0.70 & \cellcolor{orange!25}0.51 & 0.55 & 0.54 & \cellcolor{orange!25}0.63 & 0.58 & 0.59 \\
GlORIE-SLAM
 & PSNR$\uparrow$ & \cellcolor{yellow!25}23.42 & \cellcolor{yellow!25}20.66 & \cellcolor{yellow!25}20.41 & \cellcolor{yellow!25}25.23 & \cellcolor{green!25}21.28 & \cellcolor{yellow!25}23.68 & \cellcolor{yellow!25}22.45 \\
 & SSIM$\uparrow$ & \cellcolor{green!25}0.87 & \cellcolor{green!25}0.87 & \cellcolor{yellow!25}0.83 & \cellcolor{yellow!25}0.84 & \cellcolor{green!25}0.91 & \cellcolor{yellow!25}0.76 & \cellcolor{green!25}0.85 \\
 & LPIPS$\downarrow$ & \cellcolor{yellow!25}0.26 & \cellcolor{yellow!25}0.31 & \cellcolor{yellow!25}0.31 & \cellcolor{yellow!25}0.21 & \cellcolor{yellow!25}0.44 & \cellcolor{yellow!25}0.29 & \cellcolor{yellow!25}0.30 \\
Splat-SLAM
 & PSNR$\uparrow$ & \cellcolor{green!25}28.68 & \cellcolor{green!25}27.69 & \cellcolor{green!25}27.70 & \cellcolor{green!25}31.14 & \cellcolor{green!25}31.15 & \cellcolor{green!25}30.49 & \cellcolor{green!25}29.48 \\
 & SSIM$\uparrow$ & \cellcolor{yellow!25}0.83 & \cellcolor{green!25}0.87 & \cellcolor{green!25}0.86 & \cellcolor{green!25}0.87 & \cellcolor{yellow!25}0.84 & \cellcolor{green!25}0.84 & \cellcolor{green!25}0.85 \\
 & LPIPS$\downarrow$ & \cellcolor{green!25}0.19 & \cellcolor{green!25}0.15 & \cellcolor{green!25}0.18 & \cellcolor{green!25}0.15 & \cellcolor{green!25}0.23 & \cellcolor{green!25}0.19 & \cellcolor{green!25}0.18 \\
\textbf{Ours}
 & PSNR$\uparrow$ & \cellcolor{orange!25}18.91 & 18.86 & \cellcolor{orange!25}19.06 & 19.67 & x & \cellcolor{orange!25}20.37 & \cellcolor{orange!25}19.37$^\dagger$ \\
 & SSIM$\uparrow$ & \cellcolor{orange!25}0.678 & \cellcolor{orange!25}0.704 & 0.735 & 0.731 & x & \cellcolor{orange!25}0.747 & \cellcolor{orange!25}0.719$^\dagger$ \\
 & LPIPS$\downarrow$ & \cellcolor{orange!25}0.611 & 0.524 & \cellcolor{orange!25}0.531 & \cellcolor{orange!25}0.520 & x & \cellcolor{orange!25}0.502 & \cellcolor{orange!25}0.538$^\dagger$ \\
\bottomrule
\end{tabular}
\end{table}
}

\newcommand{\ScanNetResultsTable}{
\begin{table}[H]
\centering
\caption{{\bf Results on ScanNet.} Current shipped config, full sequences.}
\label{tab:scannet_results}
\tablebodyfont
\setlength{\tabcolsep}{6pt}
\renewcommand{\arraystretch}{1.15}
\begin{tabular}{|l|c|c|c|c|c|c|}
\hline
\textbf{Metric} & \textbf{\emph{0000}} & \textbf{\emph{0059}} & \textbf{\emph{0106}} & \textbf{\emph{0169}} & \textbf{\emph{0181}} & \textbf{\emph{0207}} \\
\hline
ATE Sim(3) [cm] & -- & 16.23 & -- & -- & -- & -- \\
\hline
ATE SE(3) [cm] & -- & 16.34 & -- & -- & -- & -- \\
\hline
PSNR $\uparrow$ & -- & 18.29 & -- & -- & -- & -- \\
\hline
SSIM $\uparrow$ & -- & 0.685 & -- & -- & -- & -- \\
\hline
LPIPS $\downarrow$ & -- & 0.554 & -- & -- & -- & -- \\
\hline
Depth L1 [cm] $\downarrow$ & -- & 33.16 & -- & -- & -- & -- \\
\hline
\end{tabular}
\end{table}
}

\newcommand{\SemanticScaleCorrTable}{
\begin{table}[H]
\centering
\caption{{\bf Landmark ratio tracks its intended target.} Pearson correlation (log space) between the per-keyframe landmark ratio and a ground-truth scale factor.}
\label{tab:semantic_scale_corr}
\tablebodyfont
\setlength{\tabcolsep}{6pt}
\renewcommand{\arraystretch}{1.15}
\begin{tabular}{|l|c|c|}
\hline
\textbf{Sequence} & \textbf{$r$, raw} & \textbf{$r$, frame-to-frame component ($n_{\text{lm}}\!\geq\!4$)} \\
\hline
TUM \emph{fr1/desk} & 0.43 & 0.68 \\
\hline
TUM \emph{fr1/desk2} & 0.42 & 0.56 \\
\hline
ScanNet scene \emph{0059} & 0.14 & 0.28 \\
\hline
\end{tabular}
\end{table}
}

\newcommand{\TumPsnrTable}{
\begin{table}[H]
\centering
\caption{PSNR comparison on TUM sequences}
\label{tab:psnr_results}
\tablebodyfont
\begin{tabular}{|c|c|c|c|c|c|c|}
\hline
\textbf{Method} & \textbf{\emph{fr1/desk}} & \textbf{\emph{fr2/xyz}} & \textbf{\emph{fr3/office}} & \textbf{\emph{fr1/desk2}} & \textbf{\emph{fr1/room}} & \textbf{Avg.} \\
\hline

Photo-SLAM & 20.97 & 21.07 & 19.59 & -- & -- & -- \\
\hline
MonoGS & 19.67 & 16.17 & 20.63 & 19.16 & 18.41 & 18.808 \\
\hline
GlORIE-SLAM & 20.26 & 25.62 & 21.21 & 19.09 & 18.78 & 20.992 \\
\hline
Splat-SLAM & 25.61 & 29.53 & 26.05 & 23.98 & 24.06 & 25.846 \\
\hline
Ours & -- & -- & -- & -- & -- & -- \\
\hline

\end{tabular}
\end{table}
}

\newcommand{\TumSsimTable}{
\begin{table}[H]
\centering
\caption{SSIM comparison on TUM sequences}
\label{tab:ssim_results}
\tablebodyfont
\begin{tabular}{|c|c|c|c|c|c|c|}
\hline
\textbf{Method} & \textbf{\emph{fr1/desk}} & \textbf{\emph{fr2/xyz}} & \textbf{\emph{fr3/office}} & \textbf{\emph{fr1/desk2}} & \textbf{\emph{fr1/room}} & \textbf{Avg.} \\
\hline

Photo-SLAM & 0.74 & 0.73 & 0.69 & -- & -- & -- \\
\hline
MonoGS & 0.73 & 0.72 & 0.77 & 0.66 & 0.64 & 0.704 \\
\hline
GlORIE-SLAM & 0.79 & 0.72 & 0.72 & 0.92 & 0.73 & 0.776 \\
\hline
Splat-SLAM & 0.84 & 0.90 & 0.84 & 0.81 & 0.80 & 0.838 \\
\hline
Ours & -- & -- & -- & -- & -- & -- \\
\hline

\end{tabular}
\end{table}
}

\newcommand{\TumLpipsTable}{
\begin{table}[H]
\centering
\caption{LPIPS comparison on TUM sequences}
\label{tab:lpips_results}
\tablebodyfont
\begin{tabular}{|c|c|c|c|c|c|c|}
\hline
\textbf{Method} & \textbf{\emph{fr1/desk}} & \textbf{\emph{fr2/xyz}} & \textbf{\emph{fr3/office}} & \textbf{\emph{fr1/desk2}} & \textbf{\emph{fr1/room}} & \textbf{Avg.} \\
\hline

Photo-SLAM & 0.23 & 0.17 & 0.24 & -- & -- & -- \\
\hline
MonoGS & 0.33 & 0.31 & 0.34 & 0.48 & 0.51 & 0.394 \\
\hline
GlORIE-SLAM & 0.31 & 0.09 & 0.32 & 0.38 & 0.38 & 0.296 \\
\hline
Splat-SLAM & 0.18 & 0.08 & 0.20 & 0.23 & 0.24 & 0.186 \\
\hline
Ours & -- & -- & -- & -- & -- & -- \\
\hline

\end{tabular}
\end{table}
}

\newcommand{\SemanticAblationTable}{
\begin{table}[H]
\centering
\caption{Semantic object landmarks, w/o (\textsc{Ctrl}) vs.\ w/ (\textsc{Ours}).}
\label{tab:semantic_ablation}
\tablebodyfont
\begin{tabular}{|c|c|c|c|c|}
\hline
\textbf{Sequence} & \textbf{ATE \textsc{Ctrl}} & \textbf{ATE \textsc{Ours}} & \textbf{PSNR \textsc{Ctrl}} & \textbf{PSNR \textsc{Ours}} \\
\hline
\emph{fr1/desk}       & -- & -- & -- & -- \\
\hline
\emph{fr1/desk2}      & -- & -- & -- & -- \\
\hline
\emph{fr1/room}       & -- & -- & -- & -- \\
\hline
\emph{fr2/xyz}        & -- & -- & -- & -- \\
\hline
\emph{fr3/office} & -- & -- & -- & -- \\
\hline
\end{tabular}
\end{table}
}

\newcommand{\MapItersAblationTable}{
\begin{table}[H]
\centering
\caption{Per-frame mapping iteration count, \emph{fr1/desk} (200 frames).}
\label{tab:mapiters_ablation}
\tablebodyfont
\begin{tabular}{|c|c|c|c|c|c|}
\hline
\textbf{Iterations} & \textbf{ATE [cm]} & \textbf{PSNR} & \textbf{SSIM} & \textbf{LPIPS} & \textbf{Depth RMSE [cm]} \\
\hline
30              & -- & -- & -- & -- & -- \\
\hline
60 (shipped)    & -- & -- & -- & -- & -- \\
\hline
120             & -- & -- & -- & -- & -- \\
\hline
\end{tabular}
\end{table}
}

\newcommand{\BackboneOnlyTable}{
\begin{table}[H]
\centering
\caption{{\bf Backbone-only baseline.} Unmodified SplaTAM with raw UniDepthV2 depth and no VGGT, semantics, or loop closure, full sequences, TUM RGB-D. Best \colorbox{green!25}{1st}.}
\label{tab:backbone_only}
\tablebodyfont
\setlength{\tabcolsep}{6pt}
\renewcommand{\arraystretch}{1.15}
\begin{tabular}{|l|c|c|c|}
\hline
\textbf{Configuration} & \textbf{Sim(3) ATE $\downarrow$} & \textbf{PSNR $\uparrow$} & \textbf{LPIPS $\downarrow$} \\
\hline
\emph{fr1/desk} & & & \\
\hline
Full (Ours) & 4.47 & 18.92 & 0.398 \\
\hline
SplaTAM + raw UniDepthV2 & \cellcolor{green!25}3.48 & \cellcolor{green!25}21.10 & \cellcolor{green!25}0.240 \\
\hline
\emph{fr1/desk2} & & & \\
\hline
Full (Ours) & \cellcolor{green!25}4.06 & \cellcolor{green!25}19.44 & 0.396 \\
\hline
SplaTAM + raw UniDepthV2 & 6.16 & 19.20 & \cellcolor{green!25}0.330 \\
\hline
\end{tabular}
\end{table}
}

\newcommand{\GlobalScaleTable}{
\begin{table}[H]
\centering
\caption{{\bf Is a single global metric scale sufficient?} Splat-SLAM keyframe trajectories on TUM \emph{fr1/desk2}, with a single scalar metric scale estimated three ways. ``GT-free global fit'' independently runs UniDepthV2 at every keyframe pose and fits one least-squares scalar against Splat-SLAM's own up-to-scale depth, with no ground truth anywhere in the fit. ``Ground-truth scale'' fits the single scalar that minimizes ATE directly against ground-truth trajectory, and upper-bounds what any single global scale could achieve on this sequence. Best \colorbox{green!25}{1st}.}
\label{tab:global_scale}
\tablebodyfont
\setlength{\tabcolsep}{6pt}
\renewcommand{\arraystretch}{1.15}
\begin{tabular}{|l|c|c|}
\hline
\textbf{Scale estimation} & \textbf{Scale $s$} & \textbf{ATE [cm] $\downarrow$} \\
\hline
No correction (raw, up-to-scale) & 1.0 & 64.93 \\
\hline
GT-free global fit & 2.664 & 9.76 \\
\hline
Ground-truth scale (upper bound) & 2.943 & \cellcolor{green!25}2.87 \\
\hline
Ours (frame-dependent correction) & -- & 4.06--4.87 \\
\hline
\end{tabular}
\end{table}
}

\newcommand{\SelfReinforcingBiasTable}{
\begin{table}[H]
\centering
\caption{{\bf Self-reinforcing bias check.} TUM \emph{fr1/desk2}, full sequence. A fixed, incorrect $+15\%$ bias is injected into UniDepthV2's raw depth for frames 0--80, then removed; the remaining frames are unmodified. ``Correction on'' is the shipped semantic scale correction; ``correction off'' computes and logs the same measurement without applying it to the depth target. Best \colorbox{green!25}{1st}.}
\label{tab:self_reinforcing_bias}
\tablebodyfont
\setlength{\tabcolsep}{5pt}
\renewcommand{\arraystretch}{1.05}
\begin{tabular}{lcccccc}
\toprule
Configuration
& Sim(3) ATE $\downarrow$
& SE(3) ATE $\downarrow$
& PSNR $\uparrow$
& SSIM $\uparrow$
& LPIPS $\downarrow$
& Depth RMSE $\downarrow$ \\
\midrule

Control (no injected bias)
& \cellcolor{green!25}4.40
& \cellcolor{green!25}5.30
& \cellcolor{green!25}19.25
& \cellcolor{green!25}0.715
& \cellcolor{green!25}0.399
& 23.06 \\

Bias injected, correction on
& 6.63
& 8.11
& 18.74
& 0.709
& 0.412
& \cellcolor{green!25}22.68 \\

Bias injected, correction off
& 6.15
& 7.17
& 18.89
& 0.711
& 0.401
& 23.31 \\

\bottomrule
\end{tabular}
\end{table}
}

\newcommand{\FalseLCStressTable}{
\begin{table}[H]
\centering
\caption{{\bf False loop-closure stress test.} TUM \emph{fr1/desk2}, full sequence. A fabricated loop-closure candidate between two frames with no genuine revisit (verified via ground-truth poses) is given a deliberately incorrect $+0.5$m / $+15^\circ$ relative pose. ``Gate evaluated'' passes the candidate through the normal acceptance gate; ``gate bypassed'' forces the same candidate through unconditionally. Best \colorbox{green!25}{1st}.}
\label{tab:false_lc_stress}
\tablebodyfont
\setlength{\tabcolsep}{5pt}
\renewcommand{\arraystretch}{1.05}
\begin{tabular}{lcccccc}
\toprule
Configuration
& Sim(3) ATE $\downarrow$
& SE(3) ATE $\downarrow$
& PSNR $\uparrow$
& SSIM $\uparrow$
& LPIPS $\downarrow$
& Depth RMSE $\downarrow$ \\
\midrule

Control (no injection)
& \cellcolor{green!25}8.26
& \cellcolor{green!25}11.59
& \cellcolor{green!25}17.73
& \cellcolor{green!25}0.699
& \cellcolor{green!25}0.425
& \cellcolor{green!25}24.10 \\

False closure, gate evaluated
& 16.52
& 17.88
& 16.21
& 0.638
& 0.538
& 36.03 \\

False closure, gate bypassed
& 15.43
& 16.47
& 16.71
& 0.646
& 0.519
& 33.16 \\

\bottomrule
\end{tabular}
\end{table}
}

\newcommand{\FusionMechanismDatasetsTable}{
\begin{table}[H]
\centering
\hspace*{0.03\linewidth}%
\begin{tabular}{@{}p{0.44\linewidth}@{\hspace{0.06\linewidth}}p{0.44\linewidth}@{}}
\begin{minipage}[t]{\linewidth}
\vspace{0pt}
\centering
\captionof{table}{{\bf Fusion mechanism comparison, TUM \emph{fr1/desk2}.} Full sequence, same seed and protocol across arms. Best \colorbox{green!25}{1st}, \colorbox{yellow!25}{2nd}, \colorbox{orange!25}{3rd}.}
\label{tab:fusion_mechanism_tum}
\end{minipage}
&
\begin{minipage}[t]{\linewidth}
\vspace{0pt}
\centering
\captionof{table}{{\bf Fusion mechanism comparison, ScanNet \emph{0059}.} Full sequence, same seed and protocol across arms. Best \colorbox{green!25}{1st}, \colorbox{yellow!25}{2nd}, \colorbox{orange!25}{3rd}.}
\label{tab:fusion_mechanism_scannet}
\end{minipage}
\\
\begin{minipage}[t]{\linewidth}
\vspace{0pt}
\centering
\tablebodyfont
\setlength{\tabcolsep}{3pt}
\renewcommand{\arraystretch}{1.15}
\begin{tabular}{|l|c|c|c|c|c|}
\hline
\textbf{Mechanism} & \textbf{Sim(3) ATE} & \textbf{PSNR} & \textbf{SSIM} & \textbf{LPIPS} & \textbf{Depth} \\
\hline
Online (shipped) & \cellcolor{yellow!25}4.46 & 19.15 & 0.712 & \cellcolor{yellow!25}0.398 & 14.09 \\
\hline
Hand-set fixed bands & 4.78 & 19.16 & 0.712 & 0.404 & 13.73 \\
\hline
Frozen GT-curve & 4.69 & \cellcolor{yellow!25}19.27 & \cellcolor{yellow!25}0.715 & \cellcolor{green!25}0.396 & \cellcolor{yellow!25}13.50 \\
\hline
Global scalar & \cellcolor{orange!25}4.63 & \cellcolor{orange!25}19.23 & \cellcolor{orange!25}0.714 & 0.402 & \cellcolor{orange!25}13.58 \\
\hline
Confidence-based & \cellcolor{green!25}4.30\footnotemark[1] & \cellcolor{green!25}19.30 & \cellcolor{green!25}0.716 & \cellcolor{orange!25}0.401 & \cellcolor{green!25}12.91 \\
\hline
\end{tabular}
\end{minipage}
&
\begin{minipage}[t]{\linewidth}
\vspace{0pt}
\centering
\tablebodyfont
\setlength{\tabcolsep}{3pt}
\renewcommand{\arraystretch}{1.15}
\begin{tabular}{|l|c|c|c|c|c|}
\hline
\textbf{Mechanism} & \textbf{Sim(3) ATE} & \textbf{PSNR} & \textbf{SSIM} & \textbf{LPIPS} & \textbf{Depth} \\
\hline
Online (shipped) & \cellcolor{orange!25}12.03 & \cellcolor{green!25}18.86 & \cellcolor{green!25}0.704 & \cellcolor{green!25}0.524 & \cellcolor{green!25}35.51 \\
\hline
Hand-set fixed bands & \cellcolor{yellow!25}11.76 & \cellcolor{orange!25}18.60 & \cellcolor{orange!25}0.695 & \cellcolor{yellow!25}0.533 & \cellcolor{yellow!25}38.34 \\
\hline
Frozen GT-curve & 12.30 & \cellcolor{yellow!25}18.63 & \cellcolor{yellow!25}0.697 & \cellcolor{yellow!25}0.533 & 39.32 \\
\hline
Global scalar & \cellcolor{green!25}11.68 & \cellcolor{orange!25}18.60 & \cellcolor{orange!25}0.695 & \cellcolor{orange!25}0.534 & \cellcolor{orange!25}38.57 \\
\hline
Confidence-based & 13.94 & 18.36 & 0.688 & 0.545 & 40.06 \\
\hline
\end{tabular}
\end{minipage}
\end{tabular}%
\hspace*{0.03\linewidth}
\vspace{2pt}
\footnotetext[1]{On TUM this is the single best Sim(3) ATE of the five mechanisms; on ScanNet the same mechanism gives the worst ATE of the five.}
\end{table}
}

\subsection{Experimental Setup}
\label{sec:experimental_setup}
We evaluate on TUM RGB-D \citep{Sturm12iros_tumrgbd}, ScanNet \citep{Dai2017ScanNet}, and 7-Scenes \citep{Glocker13ISMAR_realtime} using one NVIDIA RTX A6000 GPU and the MonoGS post-SLAM refinement conventions \citep{Matsuki24cvpr_gsslam}. TUM RGB-D provides motion-capture poses, ScanNet uses BundleFusion poses, and 7-Scenes contains handheld Kinect sequences with camera parameters following \citep{Li26RAL_SING3R}. We report PSNR, SSIM, and LPIPS in Appendix~\ref{sec:appendix_rendering}. Sim(3) Absolute Trajectory Error (ATE) compares against monocular baselines after global scale alignment, whereas SE(3) ATE preserves metric scale for comparison with RGB-D systems. Our metric-level poses support SE(3) evaluation without fitting another global scale. Appendix~\ref{sec:appendix_evaluation_metrics} gives the precise formulas.
\subsection{Depth Model Fusion Validation}
\label{sec:depth_complementarity}
\begin{figure}[t]
\centering
\includegraphics[width=0.6\linewidth]{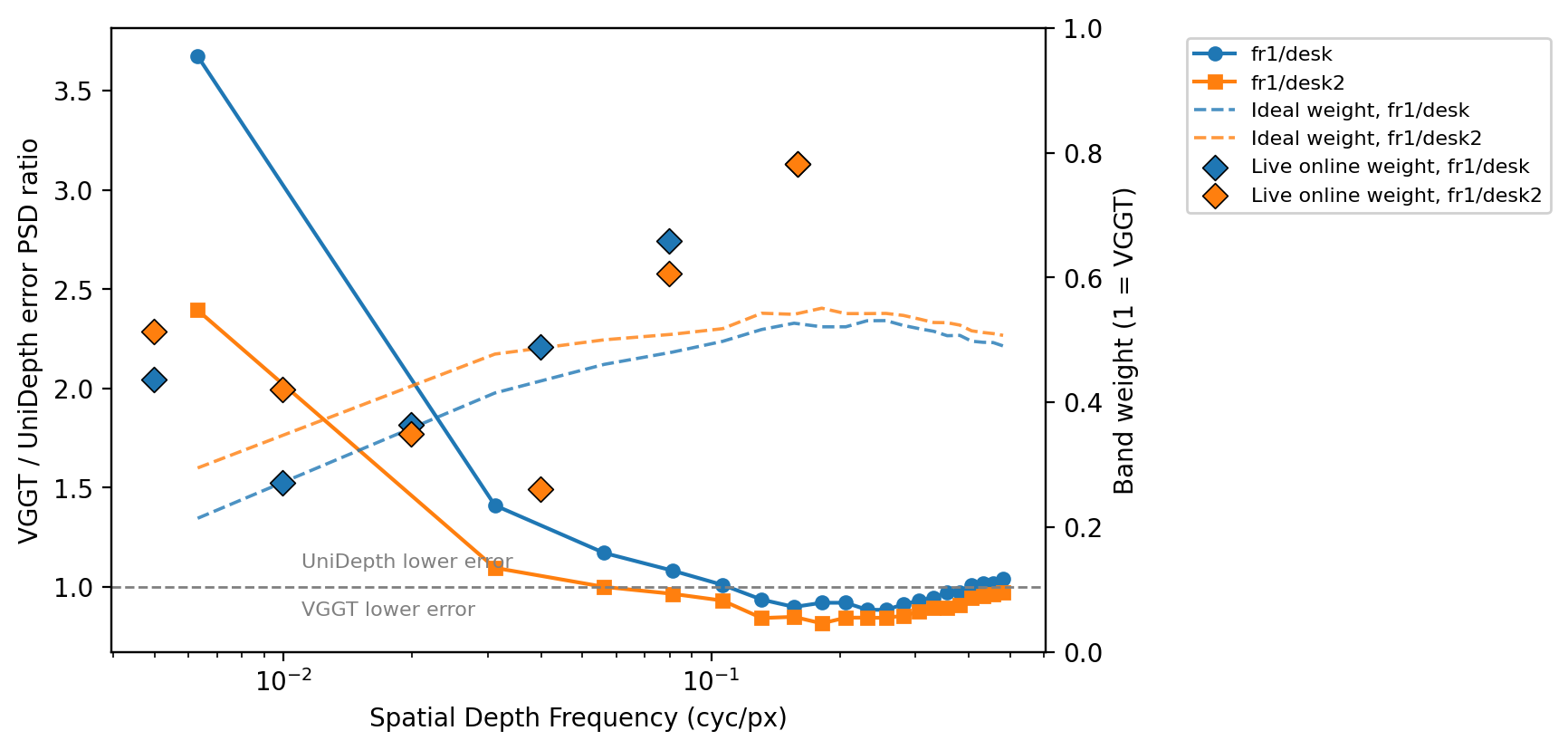}
\caption{{\bf Depth error spectral analysis between foundation models.} Ratio of VGGT's to UniDepthV2's error power spectral density vs.\ spatial depth frequency (cyc/px = cycles per pixel, a spatial-frequency measure of image sharpness), with estimated per-band weight $w_k$ overlaid (right axis, 1 = VGGT). Assuming independent model errors within each frequency band, the minimum-variance VGGT weight is $w_k^*=1/(1+r_k)$.}
\label{fig:error_spectra_ratio}
\end{figure}

Figure~\ref{fig:error_spectra_ratio} verifies that UniDepthV2 and VGGT possess complementary error structure across spatial depth-frequency bands, motivating online fusion. We measure the error power spectrum of each model against ground truth, after affine-aligning VGGT depth, and plot the resulting ratio against spatial frequency on a logarithmic axis. UniDepthV2 has lower error at lower frequencies, while VGGT is stronger at higher frequencies. The estimated per-band weights $w_k$ follow this trend, crossing $0.5$ near the region where the error ratio crosses $1$. Importantly, these weights are estimated online through Eq.~\ref{eq:spectral_band_weight} without access to ground-truth error spectra.

\subsection{Semantic Scale Correction Validation}
\label{sec:semantic_scale_validity}
Table~\ref{tab:semantic_scale_corr} reports the Pearson correlation (see Appendix~\ref{sec:appendix_evaluation_metrics}), which validates that the landmark ratio contains information about temporal scale variation. For each keyframe, we compute a ground-truth scale factor and remove slow-varying bias and drift to isolate frame-to-frame fluctuations. We then correlate this short-timescale variation with the raw landmark ratio, which tracks the frame-to-frame fluctuations at $r=0.56$--$0.68$ on TUM and $r=0.28$ on the more cluttered ScanNet scene; the correlation strengthens with more identified landmarks.
\SemanticScaleCorrTable

\subsection{Pose and Depth Estimation Performance}
\label{sec:pose_depth_estimation}
\TumAteTable
\TumDepthLOneTable
Tables~\ref{tab:tum_ate_results} and~\ref{tab:scannet_ate_results} evaluate our trajectory shape against monocular baselines after allowing global scale alignment, while SE(3) ATE retains metric scale. The fitted scale $s$ is reported in the bottom row of each table. On TUM \emph{fr1/desk}, \emph{fr2/xyz}, and \emph{fr1/desk2}, we obtain $s=0.9926,1.0046,1.02667$, showing that the estimated trajectories are already close to metric scale. In SE(3) ATE, we report the second-lowest ATE on \emph{fr1/desk2}. In ScanNet, tracking degrades substantially on scene \emph{0000}, scene \emph{0169}, and scene \emph{0207}, and fails on scene \emph{0181}. However, the fitted scale remains relatively close to one on scene \emph{0169} and scene \emph{0207}, indicating that these failures primarily arise from trajectory drift rather than global scale error.
Table~\ref{tab:tum_depth_l1} separately reports absolute metric reconstruction error directly. Full 7-Scenes results are reported in Appendix~\ref{sec:appendix_sevenscenes}.

\subsection{Loop Closure Analysis}
\label{sec:loop_closure_analysis}

\begin{figure}[t]
\centering
\includegraphics[width=0.5\linewidth]{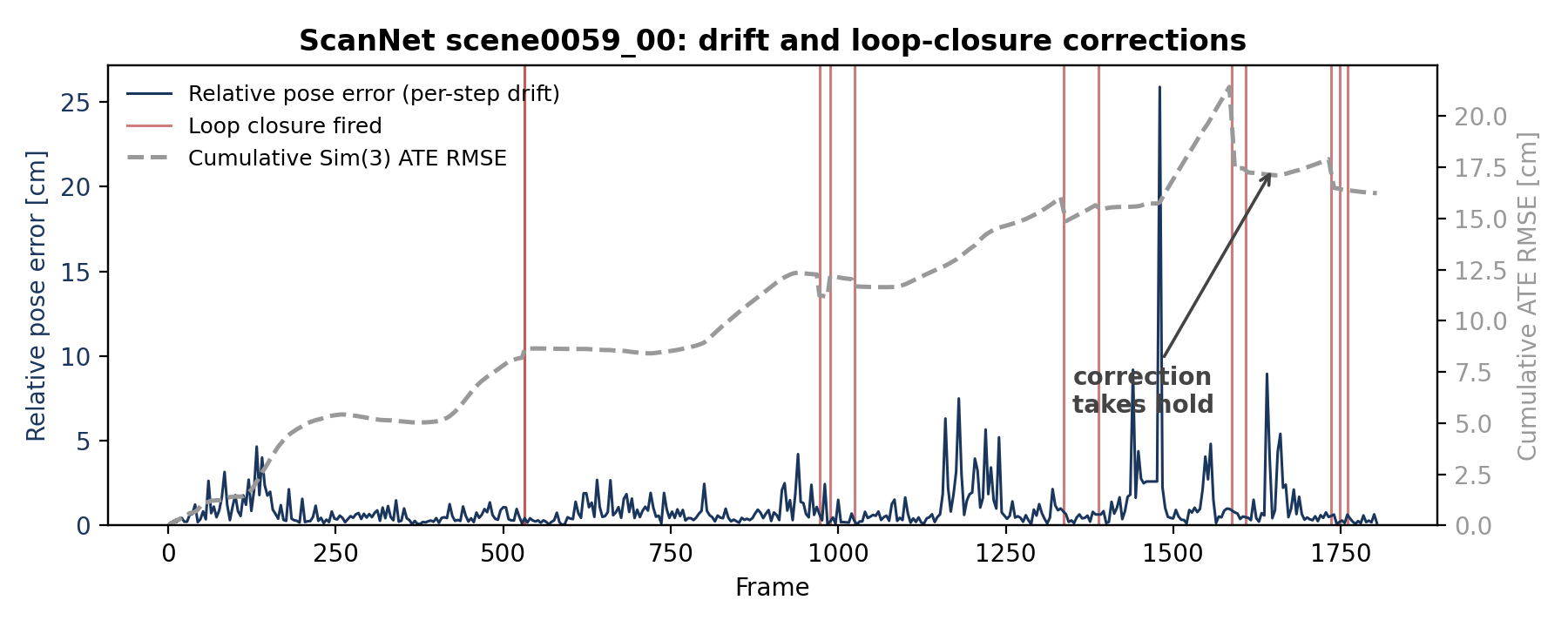}
\caption{{\bf Drift and loop-closure corrections on ScanNet scene \emph{0059}.} Per-step relative pose error (left axis) and cumulative Sim(3) ATE RMSE (right axis, dashed) against frame index. Red lines mark frames where loop closure fired.}
\label{fig:lc_drift_scannet}
\end{figure}

Figure~\ref{fig:lc_drift_scannet} reports the effects of loop closure corrections on the longer ScanNet scene \emph{0059}. Each vertical line marks a keyframe where loop closure fired. Around frame 1590, a cluster of corrections pulls the cumulative Sim(3) ATE RMSE down from a peak of roughly 21 cm; the final full-trajectory Sim(3) ATE is 16.23 cm, as reported in Table~\ref{tab:scannet_ate_results}. We report loop closure precision and recall, and a stress test of the loop-closure gate against a fabricated false candidate, in Appendix~\ref{sec:appendix_lc_table}.

\subsection{Ablations}
\label{sec:ablations}
\AblationsTable
Table~\ref{tab:ablations} isolates the contribution of each component on TUM \emph{fr1/desk2}. Removing semantic processing increases trajectory error the most, from 4.06 to 10.60 cm. Removing loop closure raises Sim(3) ATE to 7.39 cm and reduces rendering quality. Removing frequency-dependent depth fusion produces a smaller but consistent degradation. Disabling semantics removes both the scale-correction measurement and the semantic loop-closure proposals at once, so we decompose it. Scale correction alone keeps ATE close to the fully-disabled setting at 9.39 cm. Loop-closure proposals alone recover ATE to near the full-system value at 4.39 cm, so the semantic subsystem's benefit comes almost entirely from its loop-closure proposals. Fitted scale $s$ stays within 3\% of 1 for every configuration, including the row with the worst trajectory error, showing that global metric scale and trajectory shape are governed by different parts of the system.

\FusionMechanismTable
Table~\ref{tab:fusion_mechanism} compares the online per-band optimizer against a frozen band-weight curve fit offline against ground truth, shown only as an upper-bound reference since it cannot be computed at inference time. The frozen curve matches or exceeds the online optimizer on every rendering and depth metric. The online optimizer's only advantage is Sim(3) ATE. A frozen GT-fitted curve is not a deployable baseline, so the optimization's practical benefit over a fixed weighting is concentrated in trajectory accuracy, not rendering performance.

\subsection{Limitations}
Our current implementation has substantial runtimes, largely due to expensive mapping, and tracking steps (see Appendix~\ref{sec:appendix_runtime} for detailed timings). Metric accuracy inherits frame-dependent scale variation from UniDepthV2, although our failure analysis in Appendix~\ref{sec:unidepth_failure_modes} suggests that this is not the dominant source of the observed ScanNet tracking failures. VGGT can also become unreliable in scenes without sufficient parallax (see Appendix~\ref{sec:vggt_failure_modes}), which can affect both pose estimation and geometric consistency in the map. Furthermore, loop closure requires repeat observations with both semantically and geometrically verifiable matches, so our approach weakens on scenes or position revisits with few objects. Our rendering quality also trails Splat-SLAM on both TUM and ScanNet. A backbone-only baseline (unmodified SplaTAM with raw UniDepthV2, no VGGT, semantics, or loop closure) beats our full system on \emph{fr1/desk}, a comparatively static sequence, but on the more dynamic \emph{fr1/desk2} our full system pulls ahead on both Sim(3) ATE and PSNR (Appendix~\ref{sec:appendix_backbone_only}). Our added components show viability as camera motion becomes more challenging to track.

\section{Conclusion}
\label{sec:conclusion}
We introduce $S^3$-SLAM, a training-free monocular 3DGS SLAM system that estimates multi-view-consistent metric-level depth on most sequences, with a characterized failure mode when VGGT's motion estimate degenerates under weak parallax. Our results show that UniDepthV2 and VGGT possess complementary error structure across spatial depth frequencies, while semantic landmarks provide a mechanism to correct short-timescale metric-scale variation. Future work will focus on improving pose-estimation robustness and computational efficiency, extending the framework to dynamic SLAM, and applying similar methods to egocentric video.

\subsection*{AI use statement}
Generative AI tools were used to help generate mathematical formulations, iterate on algorithm design, provide feedback on hyperparameter tuning, and data analysis. We also used these tools to help with initial idea and paper section drafting. We have reviewed all AI-assisted work and take responsibility for the final content of this work, including text, claims, and artifacts produced with the aid of generative AI.
\subsection*{Reproducibility Statement}
We plan to release our code as open source on GitHub, as well as develop a project page with a summary and instructions on how to run our algorithm.

\subsubsection*{Acknowledgments}
This material is based upon work supported by the National Science Foundation Graduate Research Fellowship Program under Grant No. DGE-2039655. Any opinions, findings, and conclusions or recommendations expressed in this material are those of the author(s) and do not necessarily reflect the views of the National Science Foundation. We would also like to thank Professor Lu Gan, as well as Lingjun Zhao, and Julian Yoo for discussions related to this project.

\newpage
\appendix
\section{Appendix}
\label{sec:appendix}
\subsection{SplaTAM Conventions}
\subsubsection{3DGS Implementation}
\label{sec:appendix_3DGS}
SplaTAM \citep{Keetha24cvpr_splatam} is an RGB-D 3DGS SLAM algorithm that uses differentiable rendering via splatting for both color and depth, as shown in Eqs.~\eqref{color}--\eqref{silh}, which define the rendered color (C), depth (D), and silhouette (S), respectively, of an input pixel $\mathbf{p}$,

\begin{equation}\label{color}
C(\mathbf{p}) = \sum_{i=1}^n c_i f_i(\mathbf{p}) \prod_{j = 1}^{i - 1}\left(1 - f_j(\mathbf{p})\right)
\end{equation}
\begin{equation}\label{depth}
D(\mathbf{p}) = \sum_{i = 1}^n d_i f_i(\mathbf{p}) \prod_{j = 1}^{i - 1}\left(1 - f_j(\mathbf{p})\right),
\end{equation}
\begin{equation}\label{silh}
S(\mathbf{p}) = \sum_{i = 1}^n f_i(\mathbf{p}) \prod_{j = 1}^{i - 1}\left(1 - f_j(\mathbf{p})\right),
\end{equation}

where $c_i$ and $d_i$ are the color and camera-space depth of Gaussian $i$, respectively, and $f_i(\mathbf{p})$ is its opacity-weighted contribution at pixel $\mathbf{p}$. Under this isotropic formulation, $f_i(\mathbf{p})$ is computed by the following unnormalized Gaussian equation weighted by opacity,

\begin{equation}\label{unnormalized_gauss}
f_i(\mathbf{p}) = o_i\exp\left(-\frac{\left\lVert \mathbf{p} - \boldsymbol{\mu}^{2D}_i \right\rVert^2}{2\left(r^{2D}_i\right)^2}\right),
\end{equation}

with $o_i \in [0,1]$ being opacity, $\boldsymbol{\mu}^{2D}_i$ being the projected Gaussian center, and $r^{2D}_i$ being the projected scalar isotropic radius. The 3D center and radius are projected into pixel space by

\begin{equation}
\boldsymbol{\mu}^{2D}_i = K\frac{E_t \boldsymbol{\mu}^{3D}_i}{d_i}, \qquad
r^{2D}_i = \frac{f r_i}{d_i},
\quad
\text{where } d_i = \left(E_t\boldsymbol{\mu}^{3D}_i\right)_z,
\end{equation}

where $K$ is the camera calibration matrix, $E_t$ is the camera extrinsic matrix, $f$ is the focal length, and $d_i$ is the depth of Gaussian $i$ in camera coordinates. We use this RGB-D architecture as our base because it directly incorporates depth supervision into the mapping process.
\subsubsection{Depth Map Input and SLAM Process}
\label{appendix_map_track_dense}
Each frame is processed through three stages: tracking, mapping, and densification, operating on the set of Gaussians defined in~\ref{sec:appendix_3DGS}.

The \emph{Tracking} process estimates the current camera pose by minimizing photometric and depth re-rendering loss via Adam optimization. In our approach, the process of how this is integrated with VGGT pose estimates in the factor graph at keyframes is discussed in~\ref{sec:vggt_pose}. 

\emph{Mapping} then optimizes the existing Gaussian parameters by rendering from the current pose and minimizing depth and photometric loss, while periodically pruning low-opacity Gaussians.

\emph{Densification} inserts new Gaussians by back-projecting the current depth map if either the rendered silhouette (which holds the accumulated opacity coverage) falls below a set threshold, or if the rendered depth significantly differs from the depth target (which is obtained from raw UniDepthV2 in normal frames, or the fused VGGT-UniDepthV2 on keyframes). These conditions indicate that an occluding surface is not properly captured.

\subsection{Visual Geometry Grounded Transformers (VGGT)}
\label{appendix_vggt_formalism}
VGGT maps an image sequence to per-image predictions:

\begin{equation}\label{vggt_eqn}
f\!\left(\{I_i\}_{i=1}^{N}\right)
= \left\{\left(g_i,D_i,P_i,T_i\right)\right\}_{i=1}^{N},
\end{equation}

where $g_i \in \mathbb{R}^9$ contains the camera parameters, $D_i \in \mathbb{R}^{H \times W}$ is the depth map, $P_i \in \mathbb{R}^{3 \times H \times W}$ is the point map, and $T_i \in \mathbb{R}^{C \times H \times W}$ is a grid of features used for point tracking. 

\subsection{Factor Graphs}
\label{sec:appendix_factor_graphs}
A factor graph is a bipartite graph
$F = (\mathcal{U}, \mathcal{V}, \mathcal{E})$
\citep{Dellaert17book_factorgraphs}, where
$\phi_i \in \mathcal{U}$ represents a factor node,
$x_j \in \mathcal{V}$ represents a variable node, and $e_{ij} \in \mathcal{E}$ represents a dependency of factor $\phi_i$ on variable $x_j$. This defines the factorization of a global function
$\phi(X)$ as
\begin{equation}
    \phi(X) = \prod_i \phi_i(X_i),
\end{equation}
where $X_i \subseteq X$ denotes the subset of variables relevant to
factor $\phi_i$.

The iSAM2 algorithm~\citep{Kaess12ijrr_isam2} enables this optimization problem to be updated incrementally as new measurements and variables
are added. Internally, iSAM2 maintains a Bayes tree and recomputes only the affected cliques when new information is added. Factor-graph methods have been broadly applied to VIO and SLAM \citep{Dellaert21ar_factorGraphs}, as well as online motion planning \citep{Mukadam18CoRR_steap}.

\subsection{Semantic Tracking}
\label{sec:appendix_semantic_tracking}
This section gives the full update equations for the two filters used in Sec.~\ref{sec:semantic_scale} to estimate the semantic metric-level correction: a per-category state-space Gaussian process (SSGP), and a scalar Kalman filter that fuses its outputs across categories.

\subsubsection{Per-Category Level Estimation with an SSGP}
\label{sec:appendix_semantic_ssgp}
For each semantic category $i$, we track zero-mean Matérn-1/2 (Ornstein--Uhlenbeck) and Matérn-3/2 components. The state, measurement model, and prediction over a step $dt$ are
\begin{equation}
\mathbf{x}_t = [\text{OU}_t,\text{M32}_t,\text{M32}'_t]^\top.
\end{equation}
\begin{equation}
H = [1,1,0].
\end{equation}
\begin{equation}
y_t = H\mathbf{x}_t+\epsilon_t.
\end{equation}
\begin{equation}
R_t = \sigma^2_{\text{meas},i}.
\end{equation}
\begin{equation}
\mathbf{x}_t^- = A\mathbf{x}_{t-1}.
\end{equation}
\begin{equation}
P_t^- = AP_{t-1}A^\top+\mathbf{Q}_{\mathrm{GP}}.
\end{equation}
where $R_t$ is given by Eq.~\eqref{landmark_meas}. The matrices $A$ and $\mathbf{Q}_{\mathrm{GP}}$ are block-diagonal, with exact Ornstein--Uhlenbeck and Matérn-3/2 discretizations at step $dt$. When a measurement $y_t$ is available, the filter updates
\begin{equation}
S_t = H P_t^- H^\top+R_t.
\end{equation}
\begin{equation}
G_t = \frac{P_t^- H^\top}{S_t}.
\end{equation}
\begin{equation}
\mathbf{x}_t = \mathbf{x}_t^-+G_t(y_t-H\mathbf{x}_t^-).
\end{equation}
\begin{equation}
P_t = (I-G_tH)P_t^-(I-G_tH)^\top+G_tG_t^\top R_t;
\end{equation}
otherwise $\mathbf{x}_t = \mathbf{x}_t^-$, $P_t = P_t^-$. The per-category output consumed by the fusion filter below is
\begin{equation}
\eta_{\text{OU},i} = \exp\left([\mathbf{x}_t]_1\right).
\end{equation}
\begin{equation}
\sigma_{\text{OU},i} = \sqrt{[P_t]_{11}}.
\end{equation}

\subsubsection{Cross-Category Scale Fusion (Innovation-Gated Kalman Filter)}
\label{sec:appendix_semantic_fusion}
At each keyframe $\text{kf}_t$, categories with sufficient landmark evidence are fused using log-ratio measurements and inverse-variance weights:
\begin{equation}
\ell_i = \log \eta_{\text{OU},i}.
\end{equation}
\begin{equation}
u_i = \frac{n_{\text{used},i}}{\sigma_{\text{OU},i}^2}.
\end{equation}
\begin{equation}
z_t = \frac{\sum_i \ell_i u_i}{\sum_i u_i}.
\end{equation}
\begin{equation}
R_t = \left(\sum_i u_i\right)^{-1}.
\end{equation}
The filter predicts the prior variance under a constant-level model with process variance $q_\delta$, then forms the innovation and its variance:
\begin{equation}
P_t^- = P_{t-1}+q_\delta.
\end{equation}
\begin{equation}
\nu_t = z_t-\hat{\delta}_{t-1}.
\end{equation}
\begin{equation}
S_t = P_t^-+R_t.
\end{equation}
The measurement is accepted only if it is statistically consistent with the current state estimate under a chi-square test with one degree of freedom,
\begin{equation}
\frac{\nu_t^2}{S_t} < \chi^2_{\text{thr}}.
\end{equation}
If accepted, the state and variance are updated via the standard Kalman gain,
\begin{equation}
G_t = \frac{P_t^-}{S_t}.
\end{equation}
\begin{equation}
\hat{\delta}_t = \hat{\delta}_{t-1}+G_t\nu_t.
\end{equation}
\begin{equation}
P_t = (1-G_t)P_t^-;
\end{equation}
otherwise the measurement is rejected and the state holds, $\hat{\delta}_t = \hat{\delta}_{t-1}$, $P_t = P_t^-$. The semantic scale correction applied to the depth target is
\begin{equation}
\eta_t = \operatorname{clip}\left(\exp(\hat{\delta}_t),\ \frac{1}{1+\Delta_{\max}},\ 1+\Delta_{\max}\right).
\end{equation}

\subsubsection{Self-Reinforcing Bias Check}
\label{sec:appendix_self_reinforcing_bias}
The semantic scale correction above feeds back into the depth target the map is built from, so we test whether an incorrect correction compounds rather than self-corrects. We inject a fixed, incorrect $+15\%$ bias into UniDepthV2's raw depth for frames 0--80 of TUM \emph{fr1/desk2}, then remove it for the remainder of the sequence, and run three configurations: a control with no injected bias, the bias with correction applied as shipped, and the bias with the same measurement computed and logged but withheld from the depth target.
\SelfReinforcingBiasTable
Table~\ref{tab:self_reinforcing_bias} reports the outcome. Both bias-injected configurations are worse than the control, but the correction trace $\eta_t$ converges back toward the control's trace within roughly two to three keyframe cycles after the injected bias is removed. The configuration with correction applied and the configuration with correction withheld produce statistically indistinguishable $\eta_t$ traces throughout the run: if the correction were self-reinforcing, applying it would make the two traces diverge from each other over time; instead they track together. Applying the correction during the injected bias is slightly worse than withholding it (Sim(3) ATE 6.63 vs.\ 6.15 cm, PSNR 18.74 vs.\ 18.89), a modest, non-compounding cost from reacting to a transient error, not evidence of a self-reinforcing loop.

\subsection{Results}
\label{sec:appendix_results}
This section gives the metric formulas used in Sec.~\ref{sec:experimental_setup}, along with rendering results, the full 7-Scenes results, detailed timing analysis, depth evaluation against Splat-SLAM's protocol, and the loop closure precision/recall table.

\subsubsection{Evaluation Metrics}
\label{sec:appendix_evaluation_metrics}
For a trajectory of $N$ poses, let $\mathbf{p}_t$ and $\mathbf{p}_t^{gt}$ denote the estimated and ground-truth camera positions at frame $t$ \citep{Sturm12iros_tumrgbd}. ATE RMSE aligns the estimated trajectory to ground truth and measures the residual translation error,
\begin{equation}
\text{ATE RMSE} = \sqrt{\frac{1}{N}\sum_{t=1}^N \left\| \mathbf{p}_t^{gt} - \left(s R \mathbf{p}_t + \mathbf{t}\right) \right\|^2},
\end{equation}
where $(s, R, \mathbf{t})$ are the scale, rotation, and translation that minimize this residual. Sim(3) ATE solves for $s$ jointly with $R$ and $\mathbf{t}$. SE(3) ATE fixes $s=1$.

Following the convention in Point-SLAM \citep{Sandstrm2023PointSLAMDN}, the PSNR between a rendered image $I$ and its ground truth $I^{gt}$ is
\begin{equation}
\text{PSNR}(I, I^{gt}) = 10\log_{10}\left(\frac{1}{\text{MSE}(I, I^{gt})}\right), \qquad \text{MSE}(I, I^{gt}) = \frac{1}{HW}\sum_{p}\left\|I_p - I^{gt}_p\right\|^2,
\end{equation}
with pixel values normalized to $[0,1]$. SSIM is computed over local windows $x,y$ of $I$ and $I^{gt}$,
\begin{equation}
\text{SSIM}(x, y) = \frac{(2\mu_x\mu_y + c_1)(2\sigma_{xy} + c_2)}{(\mu_x^2 + \mu_y^2 + c_1)(\sigma_x^2 + \sigma_y^2 + c_2)},
\end{equation}
and LPIPS is the weighted distance between learned feature activations $\phi_l$ of a pretrained network at layer $l$,
\begin{equation}
\text{LPIPS}(I, I^{gt}) = \sum_l \frac{1}{H_lW_l}\sum_{p} \left\| \omega_l \odot \left(\phi_l(I)_p - \phi_l(I^{gt})_p\right) \right\|_2^2.
\end{equation}

Following the convention in \citep{Eldesokey_2020}, who correlate a predicted per-pixel confidence signal against ground-truth depth error, we validate the semantic scale correction in Sec.~\ref{sec:semantic_scale_validity} with the Pearson correlation coefficient between two scalar sequences $X$ and $Y$,
\begin{equation}
\rho_{X,Y} = \frac{\text{cov}(X, Y)}{\sigma_X \sigma_Y}.
\end{equation}

The two depth sources solve different problems and therefore fail in different ways: UniDepthV2 supplies frame-wise metric scale, while VGGT supplies multi-view geometric consistency. We analyze the ScanNet failures in Table~\ref{tab:scannet_ate_results} (scene \emph{0000}, scene \emph{0181}, and scene \emph{0207}) to determine which model is responsible, and whether our fusion and correction mechanisms can suppress the resulting error.

\begin{figure}[H]
\centering
\includegraphics[width=0.85\linewidth]{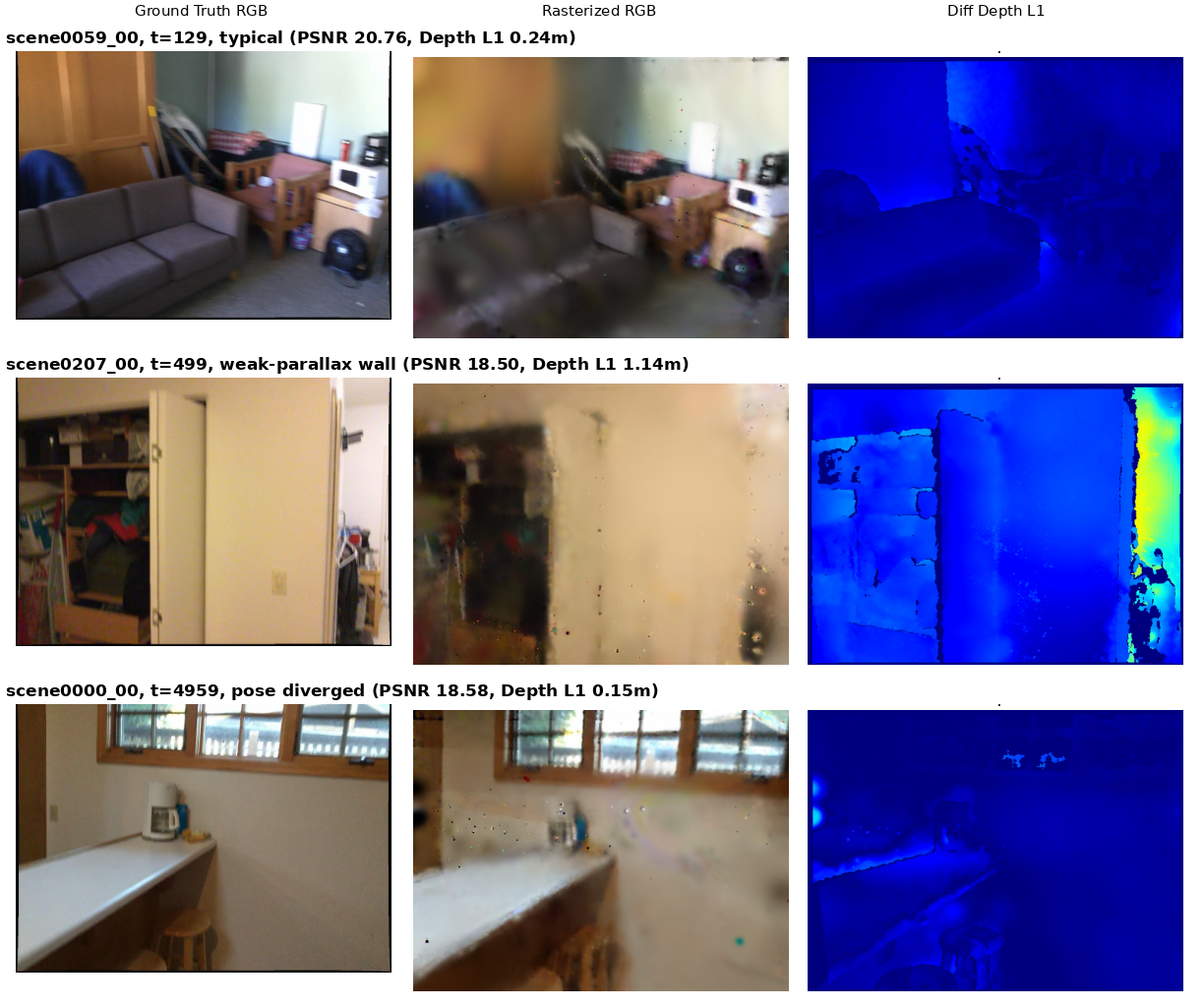}
\caption{{\bf Failure examples on ScanNet.} (Top) Ground-truth RGB, rasterized RGB, per-pixel depth error for a typical frame. (Middle) Two identified failure modes: a weak-parallax and textureless wall that causes degeneracy in VGGT's motion estimate. (Bottom) Pose has diverged but the locally rendered depth is still accurate, while the map has become degenerate.}
\label{fig:scannet_failure_modes}
\end{figure}

\subsubsection{UniDepthV2 Failure Modes}
\label{sec:unidepth_failure_modes}
We do not find UniDepthV2's per-frame depth quality to be the dominant driver of the ScanNet failures. In scene \emph{0000}'s divergence window (Fig.~\ref{fig:scannet_failure_modes}, bottom), Depth L1 stays low (0.15m) even though the trajectory has already diverged by several meters at that point. As observed in our per-frame depth analysis, UniDepthV2 exhibits affine scale variation and a near-range over-prediction bias, which our per-band depth fusion (Sec.~\ref{sec:depth_fusion}) and semantic scale correction (Sec.~\ref{sec:semantic_scale}) are built to suppress. This error would have to be substantially larger for UniDepthV2 to drive the observed failure.

\subsubsection{VGGT Failure Modes}
\label{sec:vggt_failure_modes}
VGGT's inter-frame motion estimate degenerates under weak parallax, such as when a large, nearby, textureless surface dominates the field of view and gives little multi-view triangulation signal. Fig.~\ref{fig:scannet_failure_modes} (middle) shows a representative case, with a narrow doorway opening onto a flat wall in scene \emph{0207} at $t{=}499$. The motion-derived scale ratio $r_{\mathrm{motion}}$, which should sit near 1, spikes to 12--24$\times$ its normal range at this point, later up to 142$\times$ in scene \emph{0000}, and up to 710$\times$ in scene \emph{0181}, triggering a degenerate motion detection gate in each case. In scene \emph{0181}, tracking holds steady through frame 320, then diverges sharply as the ratio climbs from 217.9$\times$ to 522.7$\times$ to 710.3$\times$ across frames 360--392. ATE rises from 5.69m to 25.6m over that window and reaches 17.23m by frame 744, with no loop closure firing to recover it. This failure is seed-dependent rather than deterministic, and a second seed under the identical configuration reaches frame 812 with ATE RMSE of only 0.48m.
\paragraph{Effect on fusion.}
The per-band depth-fusion weight (Sec.~\ref{sec:depth_fusion}) largely avoids this failure mode, since it reweights the depth target's spatial-frequency content and does not depend on the motion-derived translation scale. The pose side has no equivalent mechanism, so a degenerate VGGT motion window directly corrupts tracking. Recovery then depends on loop closure: in scene \emph{0000}, semantic loop closure fires 16 times and successfully corrects earlier drift (e.g.\ 0.79m $\rightarrow$ 0.10m near $t{\approx}$2200--2400), but stops firing once the pose corrupts past a point where our algorithm struggles to find closures. In scene \emph{0207}, loop closure never fires despite up to 87 initial loop-closure candidates per frame, because every candidate is rejected at the semantic verification gate, leaving a permanently uncorrected drift.

\subsubsection{Rendering Performance}
\label{sec:appendix_rendering}
\TumRenderingTable
Table~\ref{tab:tum_rendering_splatslam} illustrates the effect of decoupling mapping from tracking: although our Sim(3) ATE on \emph{fr1/room} is among the weakest results in Table~\ref{tab:tum_ate_results}, our PSNR and LPIPS rank second on that sequence. Note that GLORIE-SLAM \citep{zhang24arxiv_giorie} and Splat-SLAM have the same tracking, which is why GLORIE-SLAM wasn't compared earlier.

\subsubsection{Backbone-Only Baseline}
\label{sec:appendix_backbone_only}
To separate our own contribution from the capability of the UniDepthV2 backbone itself, we run unmodified SplaTAM with raw UniDepthV2 depth and no VGGT, semantics, or loop closure, on full TUM sequences \emph{fr1/desk} and \emph{fr1/desk2}.
\BackboneOnlyTable
Table~\ref{tab:backbone_only} shows a mixed result, not a uniform win for our full system. On \emph{fr1/desk}, the backbone-only baseline beats us on Sim(3) ATE, PSNR, and LPIPS. On \emph{fr1/desk2}, our full system wins Sim(3) ATE and ties PSNR, but the baseline wins LPIPS and Depth L1 by a wide margin (4.86 cm vs.\ 12.84 cm). Our added components do not establish a uniform rendering-quality advantage over this simpler baseline.

\subsubsection{Fusion Mechanism Across Datasets}
\label{sec:appendix_fusion_mechanism_datasets}
Table~\ref{tab:fusion_mechanism} in the main text compares two fusion mechanisms on TUM \emph{fr1/desk2}. Here we extend that comparison to five mechanisms on both TUM \emph{fr1/desk2} and ScanNet \emph{scene0059}: the shipped online optimizer, a hand-set fixed per-band curve, the frozen ground-truth-fitted curve, a single global scalar, and a confidence-based per-pixel blend.
\FusionMechanismDatasetsTable
Tables~\ref{tab:fusion_mechanism_tum} and~\ref{tab:fusion_mechanism_scannet} show the ranking does not transfer across datasets. On TUM, confidence-based fusion is the best mechanism on four of five metrics, including Sim(3) ATE. On ScanNet, the same confidence-based mechanism is the worst mechanism on every metric, and the shipped online optimizer wins or ties on PSNR, SSIM, LPIPS, and Depth RMSE. The online optimizer's advantage over simpler alternatives is dataset-dependent rather than a fixed margin, narrow on TUM and clearest on ScanNet.

\subsubsection{7-Scenes Pose Estimation}
\label{sec:appendix_sevenscenes}
Table~\ref{tab:sevenscenes_results} gives the full per-scene pose results on 7-Scenes using the evaluation protocol described in Sec.~\ref{sec:experimental_setup}.
\SevenScenesTable

\subsubsection{ScanNet Pose and Rendering Performance}
\label{sec:appendix_scannet}
Table~\ref{tab:scannet_rendering_results} gives rendering results on the six ScanNet scenes reported in \citep{Sandstrom25CVPRW_splatslam}, using their published baseline numbers. Scene \emph{0181} failed for our method, root-caused in Appendix~\ref{sec:vggt_failure_modes}.
\ScanNetRenderingTable

\subsubsection{Ablations on ScanNet}
\label{sec:appendix_scannet_ablations}
We repeat the main-text ablation study on ScanNet \emph{scene0059} to test whether it generalizes beyond TUM.
\ScanNetAblationsTable
Table~\ref{tab:scannet_ablations} does not replicate the TUM result cleanly. Removing loop closure is far more damaging here than on TUM (Sim(3) ATE 12.03 to 35.65 cm, roughly 3$\times$), consistent with loop closure carrying more weight on this longer trajectory. The semantic decomposition is weaker: scale correction alone has the best Sim(3) ATE of the three decomposed rows (12.12 cm), ahead of loop-closure proposals alone (12.66 cm), the opposite of the TUM ordering. Generic correspondence still underperforms the full semantic subsystem on every metric, matching TUM.

\subsubsection{Computational Overhead}
\label{sec:appendix_runtime}
\begin{figure}[H]
\centering
\includegraphics[width=0.85\linewidth]{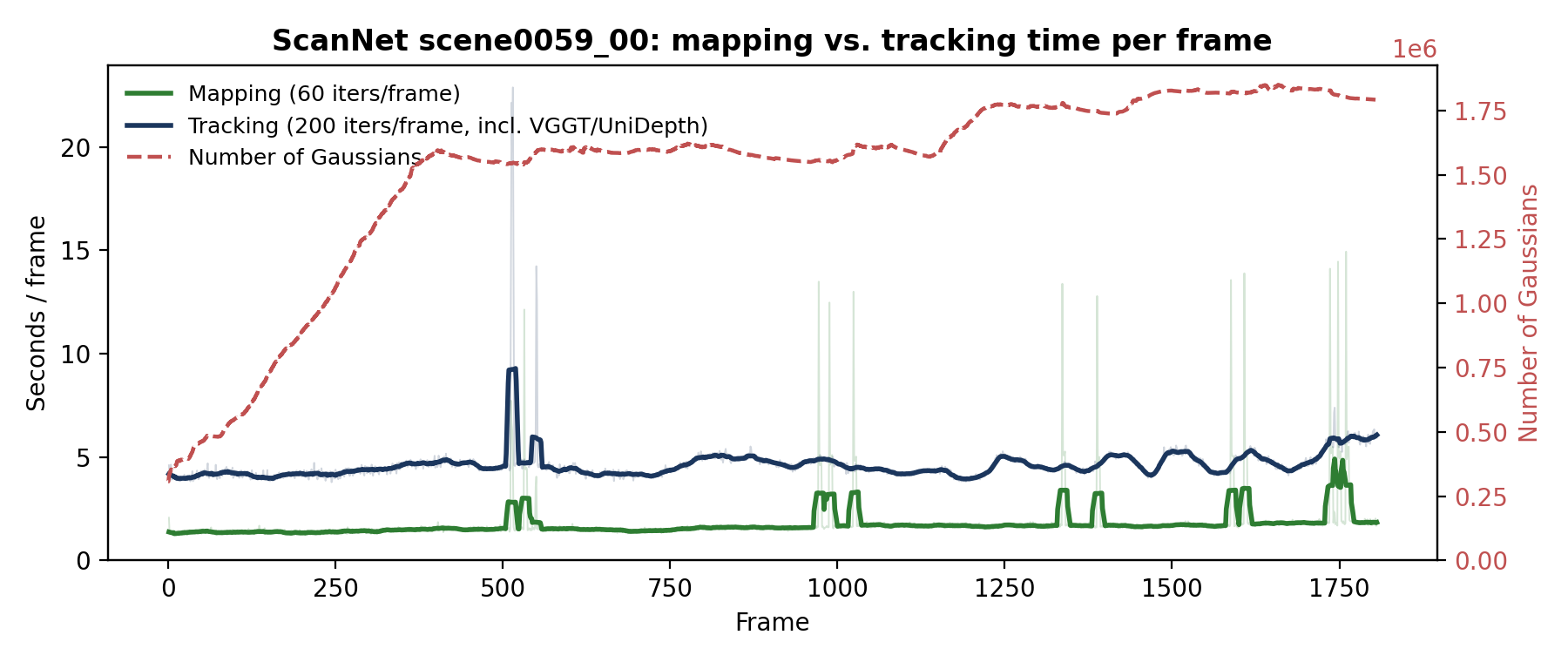}
\caption{{\bf Mapping vs. tracking time per frame on ScanNet scene \emph{0059}.} Smoothed per-frame wall-clock time for mapping and tracking (left axis) against the number of Gaussians in the map (right axis, dashed) over the course of the run.}
\label{fig:runtime_overhead_scannet}
\end{figure}

Figure~\ref{fig:runtime_overhead_scannet} breaks per-frame wall-clock time down into mapping (60 iterations/frame) and tracking (200 iterations/frame, including VGGT and UniDepthV2 inference) against the number of Gaussians in the map over the ScanNet scene \emph{0059} run. Tracking dominates at roughly 4--6 s/frame against mapping's roughly 1.5--2 s/frame. Both curves spike when loop closure fires and drift upward slightly as the map grows from 0.4M to 1.8M Gaussians over the first 400 frames.

\begin{figure}[H]
\centering
\includegraphics[width=0.7\linewidth]{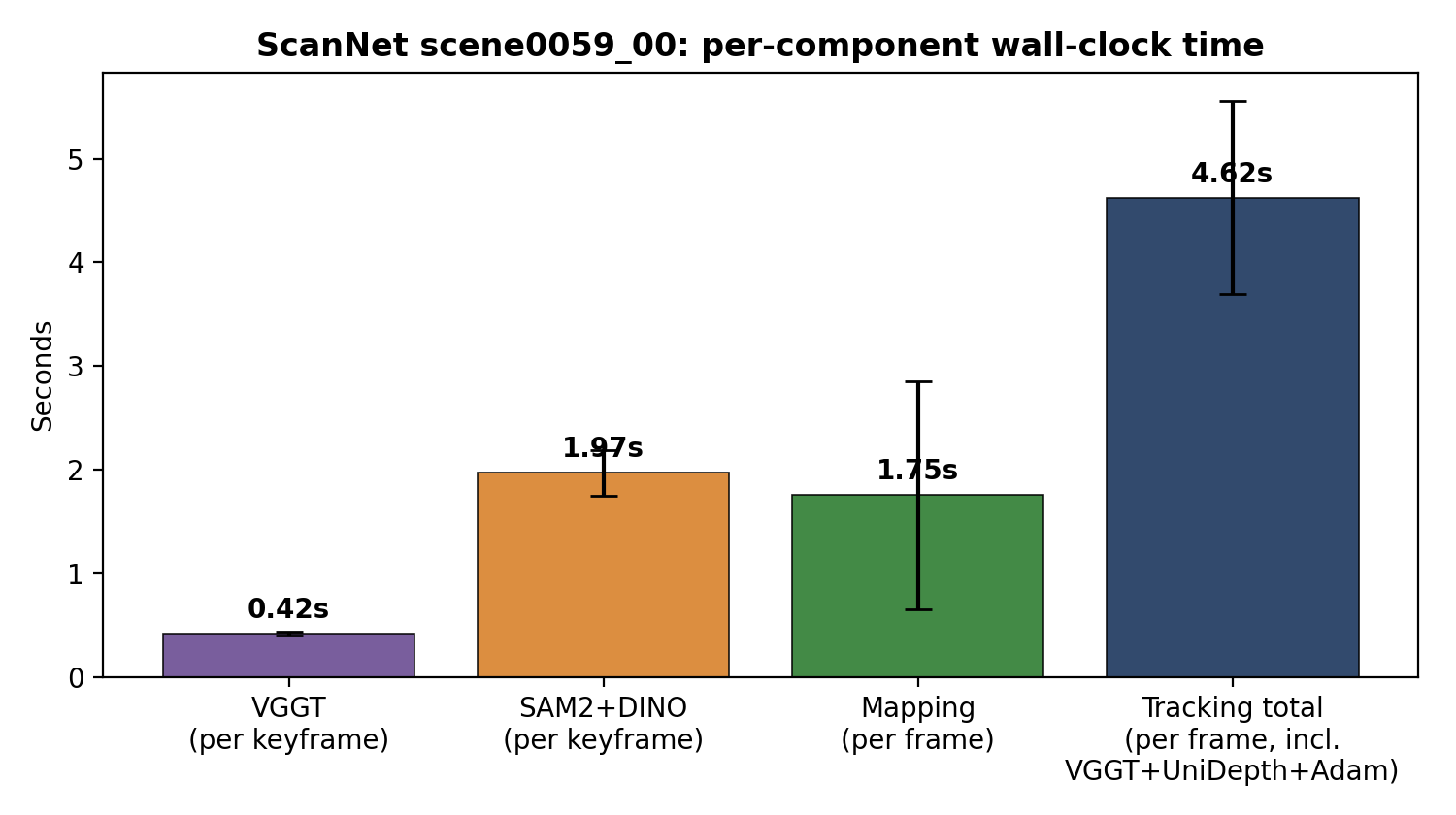}
\caption{{\bf Per-component wall-clock time on ScanNet scene \emph{0059}.} VGGT and semantic detection (Grounding DINO + SAM2) run once per keyframe, whereas mapping and the tracking total run every frame. The tracking total includes VGGT, UniDepthV2, and Adam, so the bars are not additive. Error bars show one standard deviation.}
\label{fig:runtime_breakdown_components}
\end{figure}

Figure~\ref{fig:runtime_breakdown_components} isolates components within that tracking total. The tracking total includes VGGT, UniDepthV2, and Adam and is therefore not additive with the component bars. VGGT inference on a 4-frame window costs 0.42s and is essentially constant across scenes (0.40--0.43s mean across five ScanNet scenes), since its cost is set by window size and resolution rather than scene content. Semantic detection (Grounding DINO + SAM2) costs 1.97s per keyframe. VGGT and semantic detection run only at keyframes, whereas mapping and tracking run on every frame; consequently, the latter dominate total runtime.

\subsubsection{Depth Evaluation}
\label{sec:appendix_depth_l1}
In \cite{Sandstrom25CVPRW_splatslam}, the authors do not use the ground-truth depth directly when computing Depth L1. Instead, they first compute a value $s$ that scales their Sim(3) trajectory estimate to the ground truth, and then use
\begin{equation}
    \operatorname{mean}_{t} \left[ \operatorname{mean}_{p:\ D^{GT}_{tp}>0,\ D^{rend}_{tp}>0} |sD^{rend}_{tp}-D^{GT}_{tp}| \right].
\end{equation}
Table~\ref{tab:tum_depth_comparison} reports the same evaluation after applying Splat-SLAM's Sim(3)-scaled depth protocol. Under this protocol, our average Depth L1 increases from 16.57 cm to 17.07 cm and is slightly higher than Splat-SLAM's 15.78 cm.
\begin{table}[H]
\centering
\caption{{\bf Depth L1 error [cm] on TUM RGB-D.}\ Splat-SLAM's Sim(3)-scaled protocol. Best \colorbox{green!25}{1st}, \colorbox{yellow!25}{2nd}, \colorbox{orange!25}{3rd}.}
\label{tab:tum_depth_comparison}
\setlength{\tabcolsep}{7pt}
\renewcommand{\arraystretch}{1.15}
\tablebodyfont
\begin{tabular}{|l|c|c|c|c|c|c|}
\hline
\textbf{Method}
& \textbf{\emph{fr1/desk}}
& \textbf{\emph{fr1/desk2}}
& \textbf{\emph{fr1/room}}
& \textbf{\emph{fr2/xyz}}
& \textbf{\emph{fr3/office}}
& \textbf{Avg.} \\
\hline
Original
& \cellcolor{orange!25}8.29
& \cellcolor{yellow!25}12.84
& \cellcolor{green!25}13.14
& \cellcolor{yellow!25}19.67
& \cellcolor{orange!25}28.92
& \cellcolor{yellow!25}16.57 \\
\hline
Scale+mask corrected
& \cellcolor{green!25}7.89
& \cellcolor{green!25}11.62
& \cellcolor{orange!25}18.26
& \cellcolor{orange!25}21.06
& \cellcolor{yellow!25}26.50
& \cellcolor{orange!25}17.07 \\
\hline
Splat-SLAM
& \cellcolor{yellow!25}8.05
& \cellcolor{orange!25}15.70
& \cellcolor{yellow!25}15.05
& \cellcolor{green!25}14.53
& \cellcolor{green!25}25.59
& \cellcolor{green!25}15.78 \\
\hline
\end{tabular}
\end{table}

\subsubsection{Is a Single Global Scale Sufficient?}
\label{sec:appendix_global_scale}
We test whether a single global metric scale is enough, or whether a frame-dependent correction is necessary. We clone Splat-SLAM, run it on TUM \emph{fr1/desk2}, and estimate one global metric scale from UniDepthV2 with no ground truth anywhere in the fit: independently run UniDepthV2 at every keyframe pose, and fit one least-squares scalar against Splat-SLAM's own up-to-scale depth.
\GlobalScaleTable
Table~\ref{tab:global_scale} gives the result. The ground-truth-free fit recovers most of the available scale (2.664 vs.\ the ground-truth-optimal 2.943) and cuts ATE from 64.93 cm to 9.76 cm, but this is still 2--2.3$\times$ worse than our frame-dependent correction (4.06--4.87 cm) on the same sequence. A single global scale, even fit well, is not sufficient here.

\subsubsection{Loop Closure Precision and Recall}
\label{sec:appendix_lc_table}
\LoopClosureTable

We additionally stress-test the gate with a fabricated loop-closure candidate between two frames with no genuine revisit, verified via ground-truth poses, given a deliberately incorrect $+0.5$m / $+15^\circ$ relative pose.
\FalseLCStressTable
On TUM \emph{fr1/desk2}, the gate fires on this candidate at the same confidence and drift values a genuine high-drift revisit would, and Sim(3) ATE roughly doubles for the rest of the sequence, from 8.26 to 16.52 cm, with no later closure recovering it (Table~\ref{tab:false_lc_stress}). The submap-rigid correction bounds the damage to the affected submap at the moment of the false closure, but later genuine closures operate against the now-incorrect baseline rather than correcting it, so the degradation persists.

\end{document}